\documentclass[sigconf]{acmart}

\AtBeginDocument{%
  }

\copyrightyear{2024} 
\acmYear{2024} 
\setcopyright{acmlicensed}\acmConference[ICMI '24]{INTERNATIONAL CONFERENCE ON MULTIMODAL INTERACTION}{November 4--8, 2024}{San Jose, Costa Rica}
\acmBooktitle{INTERNATIONAL CONFERENCE ON MULTIMODAL INTERACTION (ICMI '24), November 4--8, 2024, San Jose, Costa Rica}
\acmDOI{10.1145/3678957.3685735}
\acmISBN{979-8-4007-0462-8/24/11}

\usepackage{subfigure}
\usepackage{amsmath, amsthm, amsfonts,a mscd} 
\usepackage{bbm}
\usepackage{eucal} 
\usepackage{mathtools}
\usepackage{float}
\usepackage{dblfloatfix}
\usepackage{afterpage}

\begin{document}

\title{Prosody as a Teaching Signal for Agent Learning: Exploratory Studies and Algorithmic Implications}

\author{Matilda Knierim}
\authornote{Both authors contributed equally to this research.}
\affiliation{%
  \institution{Vrije Universiteit}
  \city{Amsterdam}
  \country{Netherlands}
}

\author{Sahil Jain}
\authornotemark[1]
\affiliation{%
  \institution{Sony AI}
  \city{Frisco}
  \state{Texas}
  \country{USA}}

\author{Murat Han Aydoğan}
\affiliation{%
  \institution{Koç University}
  \city{Istanbul}
  \country{Turkey}
}

\author{Kenneth Mitra}
\affiliation{%
 \institution{University of Texas at Austin}
 \city{Austin}
 \state{Texas}
 \country{USA}}

\author{Kush Desai}
\affiliation{%
  \institution{University of Texas at Austin}
  \city{Austin}
  \state{Texas}
  \country{USA}}

\author{Akanksha Saran}
\authornote{Equal advising.}
\affiliation{%
  \institution{Sony AI}
  \city{San Francisco}
  \state{California}
  \country{USA}
  }
\email{akanksha.saran@sony.com}

\author{Kim Baraka}
\authornotemark[2]
\affiliation{%
  \institution{Vrije Universiteit}
  \city{Amsterdam}
  \country{Netherlands}}
\email{k.baraka@vu.nl}

\renewcommand{\shortauthors}{Knierim, Jain et al.}

\begin{abstract}
  Agent learning from human interaction often relies on explicit signals, but implicit social cues, such as prosody in speech, could provide valuable information for more effective learning. This paper advocates for the integration of prosody as a teaching signal to enhance agent learning from human teachers. Through two exploratory studies—one examining voice feedback in an interactive reinforcement learning setup and the other analyzing restricted audio from human demonstrations in three Atari games—we demonstrate that prosody carries significant information about task dynamics. Our findings suggest that prosodic features, when coupled with explicit feedback, can enhance reinforcement learning outcomes. Moreover, we propose guidelines for prosody-sensitive algorithm design and discuss insights into teaching behavior. Our work underscores the potential of leveraging prosody as an implicit signal for more efficient agent learning, thus advancing human-agent interaction paradigms.
\end{abstract}


\received{16 August 2024}
\maketitle

\section{Introduction}
Recent years have witnessed a surge in research focused on how agents can learn from human interactions, predominantly concentrating on clear and overt cues such as natural language feedback~\cite{lin2020review,liang2024learning, goyal2022using}. Nonetheless, human communication is inherently complex, infused with a variety of subtle and implicit signals that, we hypothesize, could significantly enhance the agent's learning process. The field of interactive learning with multi-modal human cues ~\cite{lockerd2004tutelage,lin2020review} has started leveraging implicit signals such as clicker-based feedback (perfect and imperfect)~\cite{faulkner2020interactive,zhang2021recent,li2019human}, eye movements ~\cite{saran2018human, saran2020efficiently, saran2020understanding, zhang2020human}, facial expressions~\cite{cui2020empathic,lin2020human}, gestures~\cite{visi2020towards,lin2020human}, haptic feedback~\cite{cruz2018multi,chen2021reinforcement}, and object and environmental sounds~\cite{zhang2019cutting,gandhi2019swoosh,dean2020see,aytar2018playing}. 

In this emerging research landscape, one modality stands out as clearly underexplored. \textit{Prosody}, which involves various acoustic properties of speech such as tone, pitch, rhythm, and intonation, plays a critical role in human-human and human-animal interactions, where it serves as a key vehicle for conveying emotions, intentions, and expectations~\cite{spinelli2017does,gergely2023dog}. For example, a rising intonation at the end of a sentence can distinguish an assertion from a question, or varying stress on words can alter the meaning of a phrase, demonstrating the nuanced role of prosody in communication. Despite its clear importance, prosody has not been extensively studied as a teaching signal within the realm of agent learning. This paper seeks to bridge this gap by delving into the potential of prosody to act as an informative signal for agents to learn from. By examining how prosody can aid in the interpretation and understanding of verbal instructions, we aim to underscore its value not only in facilitating agent learning but also in enriching our general understanding of human interaction, thereby contributing to more nuanced and effective communication models.

This paper considers the setting where a human who is an expert at the task uses speech as a teaching modality to teach an agent a near-optimal policy. We specifically consider two scenarios: one where a reinforcement learning (RL) agent needs to learn a policy from binary speech feedback (Sec.~\ref{intRL-study}), and the other where an imitation learning agent learns from speech-augmented demonstrations, i.e., the expert talks while providing full demonstrated trajectories (Sec.~\ref{demo-study}). For both studies, we focused only on ``Yes''/``No'' feedback. While richer evaluative feedback could be employed in practice and converted to binary feedback, we opted for a more controlled setting to reduce confounding factors related to variations in speech content. Our contributions are as follows:

\begin{itemize}
\item A user study ($N=28$) demonstrating links between prosodic features and task-related features in a Wizard-of-Oz interactive RL setting (grid-world task).
\item A pilot study ($30$ mins of audio from a single teacher) suggesting a similar role of prosody in speech-augmented demonstrations for three Atari games.
\item Preliminary evidence that incorporating prosody in interactive learning algorithms can improve learning performance.
\item Documented open source data collection pipelines, including a visualization replay tool, to facilitate similar data collection efforts by other researchers.\footnote{\label{codebase_footnote} Link to code repository for both experiments: \href{https://github.com/sahiljain11/audio_rl}{https://github.com/sahiljain11/audio\textunderscore rl}}
\end{itemize}

\section{Related Work}
We provide a brief overview of work on use of human speech to augment two learning paradigms evaluated in this work: (a) interactive RL (Sec.~\ref{sec:related_int_rl}), and (b) learning from demonstration (Sec.~\ref{sec:related_lfd}).

\subsection{Speech-assisted Interactive RL}
\label{sec:related_int_rl}
Some prior works which leverage human feedback during reinforcement learning tasks \cite{zhang2019leveraging} do so via voice
 \cite{tenorio2010dynamic,kim2007learning,kim2009people,
krening2018newtonian,krening2016learning}. 
Tenorio et al.~\cite{tenorio2010dynamic} perform reward shaping using SARSA \cite{rummery1994line, singh1996reinforcement, sutton1998introduction} with human voice. Under their setup, the voice-based feedback is provided as the robot is executing the task. 
However, 
rewards are predefined for certain words in a vocabulary of $250$ words such as +50 for ``excellent'', $-50$ for ``terrible'' etc., and no prosodic information is used. 
Krenig et al.~\cite{krening2018newtonian} train RL agents with action advice in the form of human voice, such that a set of predefined words directly map to an underlying action from a discrete action set. Krenig et al.~\cite{krening2016learning} use sentiment analysis to filter explanations into advice of what to do and warnings of
what to avoid. 
Nicolescu et al. \cite{nicolescu2003natural} demonstrated the role of verbal cues both during demonstrations and as feedback from the human teacher during the agent's learning process, to facilitate learning of navigation behaviors on a mobile robot, with a limited vocabulary of words to indicate relevant parts of the workspace or actions that a robot must execute. 
However, all of these prior works focus on the spoken words and do not leverage prosody from human speech---an informative and rich signal of human intent which has the potential to further enhance learning~\cite{saran2022understanding}.

Kim et al. \cite{kim2007learning} use affective human speech feedback over $25$ msec audio snippets to improve a social waving behavior using Q learning. 
They use three prosody features (total band energy, variance of log-magnitude-spectrum, variance of log-spectral-energy) to learn the wave that optimally satisfies a human tutor. We build on this work to further understand how prosodic features relate to RL-specific features, such as the reward and advantage function, with the goal to inform the design of future algorithms with an underlying RL formulation (e.g., interactive RL, imitation learning) which can be more sample efficient by leveraging prosodic information.

\subsection{Speech-assisted Learning from Demonstration}
\label{sec:related_lfd}

Prior research in learning from demonstrations (LfD) \citep{argall2009survey, osa2018algorithmic} has utilized human speech signals accompanying demonstrations. 
\citet{nicolescu2003natural} demonstrate the role of verbal cues both during demonstration and feedback from the human teacher to facilitate learning or sequential arrangement of navigation behaviors on a mobile robot. However, they use a fixed vocabulary of words that a demonstrator can use to indicate relevant parts of the workspace or actions that a robot must execute. 
\citet{pardowitz2007incremental} show how vocal comments with a demonstration can augment subtask similarity and learning the task model (task precedence graph) for a simple table setting task. They use a fixed set of seven vocal comments which are mapped one-to-one with features relevant to the task. \citet{rybski2007interactive} learn a planning model from human demonstrations and dialog for a mobile robot. They map human utterances to match them against a  set of predefined natural language commands and manually ground them to locations on a map which the robot has access to for learning and executing navigation behaviors.
In our work, however, we study how prosody in speech relates to RL-specific features and accordingly leverage prosody to aid two interactive learning paradigms.

Recently, ~\citet{saran2022understanding} characterize unrestricted speech from human teachers demonstrating multi-step manipulation tasks to a situated robot. They report differentiating properties of speech in terms of duration and expressiveness, highlighting that human prosody carries rich information useful for enhancing LfD. In our work, we propose a novel algorithm that can leverage prosody for LfD and validate it in three simulated game environments.

\section{Study 1: Voice Feedback in Interactive RL}
\label{intRL-study}
This study involved analyzing voice feedback provided by human trainers in an interactive RL (intRL) setup. We investigated whether prosodic features in ``Yes''/``No'' feedback correlated with task performance metrics, thus shedding light on the role of prosody in implicit teaching signals.

\subsection{Mixed-participant Wizard-of-Oz setup}
\label{sec:woz}
\begin{figure*}[hpbt]
\centering
\includegraphics[width=.9\textwidth]{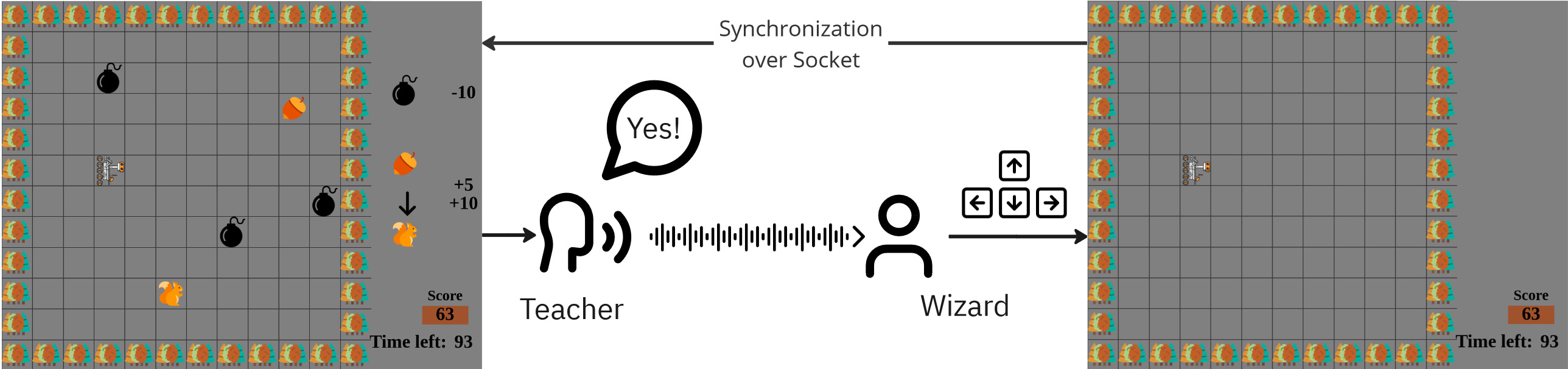}
\caption{Mixed-participant remote Wizard-of-Oz setup with the teacher view (left) and Wizard view (right).}
\vspace{-2mm}
\label{fig:fig1}
\end{figure*}

In order to develop an algorithm that leverages implicit information in prosody, we need to first understand how people use prosody as a teaching signal. On the other hand, in order to understand prosodic behavior in this context, we need to have an algorithm that incorporates prosody in its learning, which we don't have yet. To solve this paradox, we opted for a mixed participant Wizard-of-Oz (WoZ) approach where one participant plays the role of a teacher, and the other participant with no information about the task plays the role of a Wizard. This setup makes sure that the teacher audio we get is as close a possible to what we would expect in our target context. This approach is superior to using a baseline algorithm for the agent (e.g., intRL based on speech only) for two reasons. First, we expect the teacher to adapt to the learner, thereby potentially suppressing their prosodic signals (which was confirmed in some early pilots we ran with fixed agent trajectories). Second, the wizard's keystrokes provide us with valuable data that can be used in future research to better understand local and global interpretations of teacher feedback, independently of how well the teacher is able to teach.

Our contributed web-based WoZ interface is shown in Fig.~\ref{fig:fig1}. While the teacher sees the full environment and provides online verbal feedback to the agent in real-time, the Wizard is only shown a sanitized view of the environment that solely shows the grid. The Wizard also receives the teacher audio in real-time (streamed through Twilio, a secure web service) and needs to control the agent through keystrokes in response to current and past feedback from the teacher. The two environments are synchronized over web sockets to ensure consistent agent positions on both interfaces, and data is automatically and securely stored on a cloud bucket. The code for this pipeline, including a data visualization replay tool, is made open-source in order to facilitate further research in the interactive learning community\footnote{\href{https://github.com/sahiljain11/audio_rl}{https://github.com/sahiljain11/audio\textunderscore rl}}.

\subsection{Study Design}

\begin{figure*}
\minipage{0.32\textwidth}
\centering
  \includegraphics[width=.8\linewidth]{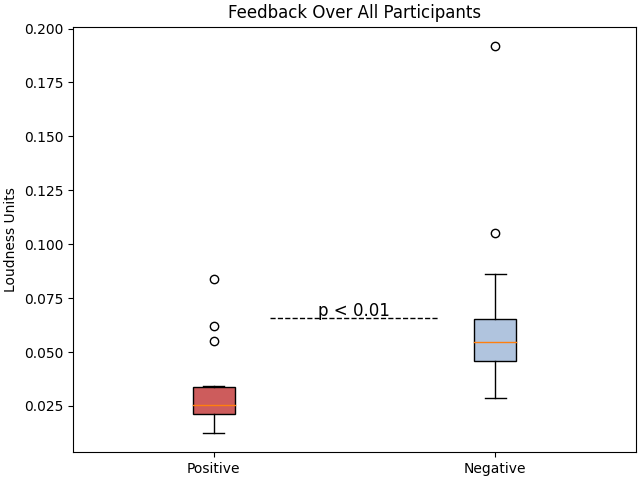}
  \caption{Yes/No balance for loudness}\label{fig:comparison_loudness}
\endminipage\hfill
\minipage{0.32\textwidth}
\centering
  \includegraphics[width=.8\linewidth]{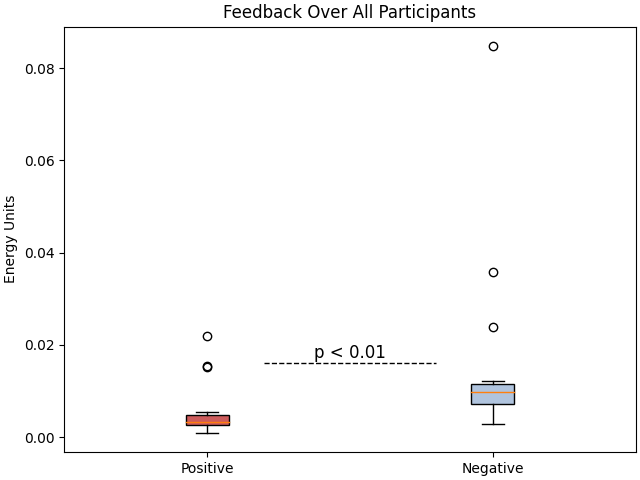}
  \caption{Yes/No balance for energy}\label{fig:comparison_energy}
\endminipage\hfill
\minipage{0.32\textwidth}%
\centering
  \includegraphics[width=.8\linewidth]{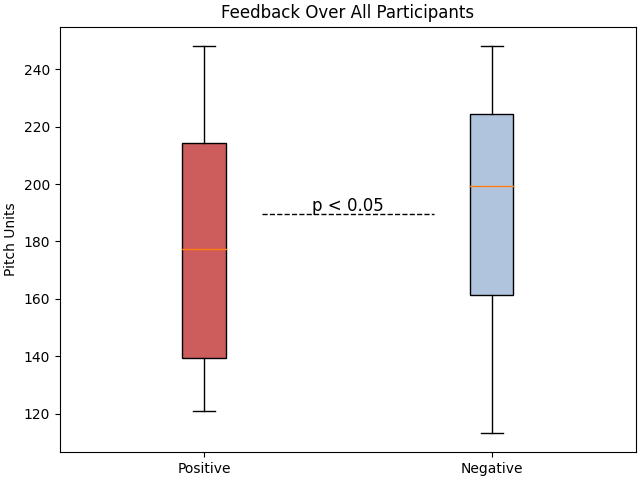}
  \caption{Yes/No balance for pitch}\label{fig:comparison_pitch}
\endminipage
\end{figure*}

\subsubsection{Participants.}
We recruited $28$ participants to pair up in a total of $14$ sessions.
Most of the participants were students, except for an operations manager, a psychologist, a teacher, and a student assistant for teacher professionalization training. The mean age of the participants was $24$ years old, with $16$ identifying as female, seven as male, and none as other genders.

\subsubsection{Experimental Setup.} Our experiment used our mixed- participant WoZ setup on a Robotaxi environment~\cite{cui2020empathic} in which the agent had to pick up a nut and deliver it to a squirrel while avoiding three bombs (see Fig.\ref{fig:fig1}). Even though the agent was human-controlled in this experiment, in order to quantify task-related metrics, we modeled the underlying task as a Markov Decision Process (MDP) where discrete states represented the agent's position, actions up/ down/ left/ right were available to the agent, with deterministic state transitions and rewards at special states (shown in Fig.~\ref{fig:fig1}) in addition to a cost-to-live of 1 per time step. After some piloting iterations, we chose a timestep duration of $1.25$ secs which made it not too boring nor too challenging for participants, while mitigating network delays (which were on average below that number).

The map was created with wall borders around the playable area. The robot location was initialized randomly at the beginning of the game, with a random initial travel direction that ensured that it could move at least three spaces in its starting direction without hitting a wall. Game elements were placed with the following constraints: bombs within at least four spaces of the robot's initial direction of travel and a Manhattan distance of at least $4$ between all pairs of elements. The wizard remotely controlled the robot with arrow keys on the keyboard. In the absence of wizard input for at least one timestep, the robot started exploring the map randomly, mimicking exploration/exploitation phases typical of RL algorithms~\cite{sutton1998introduction}. The experiment consisted of one practice round (until the goal was reached) and three game rounds for analysis. 

\vspace{-1mm}
\subsubsection{Procedure.} The study was approved by university ethical committees of two of our affiliations. Pairs of participants were appointed simultaneously for the study and welcomed by the examiner separately in either the teacher or the wizard role. All participants received a relevant consent form prior to the session and were compensated with a $10$ EUR/USD gift card for participating in the study. All experimental
sessions lasted between 15-30 minutes. Some participants optionally consented to make their data (including audio recordings) publicly available. One such sample session is included as a video recording in supplementary material.

\textit{Teacher} -- To establish a baseline of the teacher's voice prosody, the participant read a small paragraph given to them at the beginning of the study session ($\sim30$ seconds to read). During the rest of the study, the teacher was only allowed to use the words `yes' and `no' as feedback to the agent. To elicit richer prosody variations, we instructed the teachers to say these words as if they would to a $2$ year old child. Before the study session, the teacher was told that the agent would be listening to their voice, including ``how'' they spoke, and acting accordingly. In reality, another participant in the role of wizard was controlling the agent in response to the teacher's verbal feedback (Fig.~\ref{fig:fig1}). 
    
\textit{Wizard} --- The wizard was briefed on the setup of the study. They had, however, no knowledge about what the agent's task entailed or where the special states were located. Their keyboard interactions were recorded alongside other details about the game such as immediate rewards, timestamps, movements of the agent, and score. After the experiment, the wizard was asked not to talk about the experiment procedure with other potential participants.

\subsubsection{Measures.} In this paper, we only focus on measures related to the teacher's data, namely:

    \textbf{\textit{Prosodic Features:}} We computed several acoustic features over the detected utterances to characterize prosody from human verbal feedback.  We studied five different features to capture prosody -- \textbf{utterance duration}, \textbf{utterance repetition}, \textbf{pitch}, \textbf{energy}, and \textbf{loudness}. These acoustic features have been shown to enhance semantic parsing \cite{tran2017parsing}, understand speech recognition failures in dialogue systems \cite{hirschberg2004prosodic}, and are widely used for applications to human-robot interaction~\cite{kim2009people,short2018detecting} and speech recognition~\cite{yu2016automatic} communities. These features were extracted from the audio recording of the experiment sessions.

    \textbf{\textit{MDP Features:}} As in most reinforcement learning settings, we model the learning task as a Markov Decision Process (MDP). The Q-value $Q(s,a)$ represents the expected utility of action $a$ at state $s$ if the agent follows the optimal policy. From it, the \textbf{advantage function} is calculated as: $A(s,a) = Q(s,a) - V(s)$. Recent work by Cui et al.~\cite{cui2020empathic} states that advantage might be a better task statistic to consider than reward for analyzing social signals (interpreted as evaluative feedback) in interactive reinforcement learning setups. Our hypothesis is that the advantage function, as a measure of relative performance of a given action at a given state, given the overall optimal policy, would significantly correlate with our prosodic features, with potential differences across different teachers based on their expressivity levels.

    \textbf{\textit{Teaching performance:}} We used agent/wizard performance as a proxy for teaching performance by counting the number of timesteps to reach the goal. Since the environment was differently set up for each participant, we normalized the performance by dividing the absolute steps of the robot by the optimal number of steps.
    
    \textbf{\textit{Demographic measures:}} We collected a number of demographic measures, including experience training/teaching others in a professional setting (5-point Likert item), and having a pet. The complete survey can be accessed online. \footnote{Survey URL: \href{https://tinyurl.com/SurveyAudioRL}{https://tinyurl.com/SurveyAudioRL}}

\subsubsection{Data processing}
Transcriptions of the recordings were created using Google Cloud’s Speech-To-Text~\cite{googlestt}. The transcriptions were used to filter out silent parts and speech other than ``yes'' and ``no'' (which mostly consisted of non-verbal sounds). 
We encoded the positive feedback "yes" to $1$ and the negative feedback "no" to $-1$. The prosodic feature values for each feedback utterance were calculated as follows. \textit{Utterance duration} was estimated by calculating the time difference between the start and end timestamps of the word transcription. \textit{Utterance repetition} was identified by analyzing word chunks within the transcription and assessing if a word was repeated consecutively as a flag $\in \{0, 1\}$. If a word was part of a repetitive word chunk, each word of the chunk got the same label. For the other prosodic features, we vectorized audio recordings with the python library Librosa \cite{librosa}. To estimate the \textit{pitch}, we employed the Librosa Yin function \cite{librosa}, which provides a fundamental frequency estimation. \textit{Energy} and \textit{loudness} were computed as $energy = \frac{\sum_{1}^{N}(x_{i})^2}{N}$ and $loudness = \frac{\sum_{1}^{N}abs(x_{i})}{N}$, where $x_1, ...x_N$ denotes the acoustic signal \cite{kim}. The prosodic feature \textit{loudness} here corresponds to the sound pressure recorded by the microphone. We chose this representation since it resembles how the agent receives the audio signal, as opposed to other measures or loudness (e.g., in phon) which factor in subjective human hearing. The range of all prosodic feature values except repetition was $x \in [0, \infty)$. We combined the binary speech mapping with the prosodic features by multiplying them with the feedback values. Consequently, the sign of the prosodic values gets flipped for the negative feedback.

Due to the remote nature of the experimental setup, some timestamps would occasionally be skipped by the logger. However, in most cases it was easy to interpolate the robot's location and actions based on the previous and following timestamp for the advantage function calculation. Any missing data that was more than a couple timestamps was omitted from the analysis. Four out of 14 sessions were intact and did not need any interpolation.
\vspace{-5mm}

\subsection{Results}

\subsubsection{Positive/negative bias}
\label{sec:results_intrl_pos_neg}

In line with previous work~\cite{thomaz}, we observed significantly more positive than negative feedback for five out of 14 trainers (\textit{p}-values $\leq 0.003$ on a Bonferroni corrected chi-square goodness of fit test). The other participants did not show statistical significance. A population level  chi-square goodness of fit test showed that there was significantly more positive than negative feedback over all participants (\textit{p} $\leq 0.001$). Therefore, we conclude that the usage of positive and negative feedback was unbalanced and more positive feedback was used.

Our results also showed that the extent to which prosody features (loudness, pitch, and energy) were used is higher for negative than positive feedback. We did not observe the same effect for duration and repetition features. We ran t-test analyses for each prosodic feature and each participant to compare the prosodic expressions. The results can be found in Fig.~\ref{fig:comparison_loudness}, \ref{fig:comparison_energy}, \ref{fig:comparison_pitch}. The plots show the aggregated values for each prosodic feature over all participants. The aggregated \textit{p}-values were determined with a Bonferroni correction. When we look at each prosodic feature for different participants, 
energy and loudness have 13 out of 14 significant differences between the prosodic features of the feedback word "yes" and "no". This highlights that participants spoke louder and with higher energy when they gave negative feedback to the agent. Pitch has a less clear distinction between positive and negative feedback. Seven out of the 14 trainers had a significantly different pitch value when saying the word "no" than "yes". 

\subsubsection{Link to MDP metrics}

For each participant, advantage values were correlated with the corresponding word's prosodic feature. Spearman's rank correlation coefficient was used with a Bonferroni correction to determine the relation between the variables since the advantage function is not normally distributed. For the ``repetition'' feature, a point-biserial correlation was used.

For loudness and energy, three out of $14$ correlations were significant, with a correlation coefficient of $0.15$ and $0.25$. This suggests that people give positive feedback with higher energy and loudness if the taken action was the best possible one. In contrast, a sub-optimal action is associated with negative feedback rich in energy and loudness. One correlation had a negative coefficient of $-0.25$, suggesting that that person gave prosody-rich positive feedback when actions were suboptimal and prosody-rich negative feedback when the action was optimal. In order to compare the correlation results at the population level, we conducted a repeated measure correlation. The result showed that the correlation did not hold at the population level with a \textit{p}-value = $0.23$.

\begin{figure}
    \centering
    \includegraphics[width=0.3\textwidth]{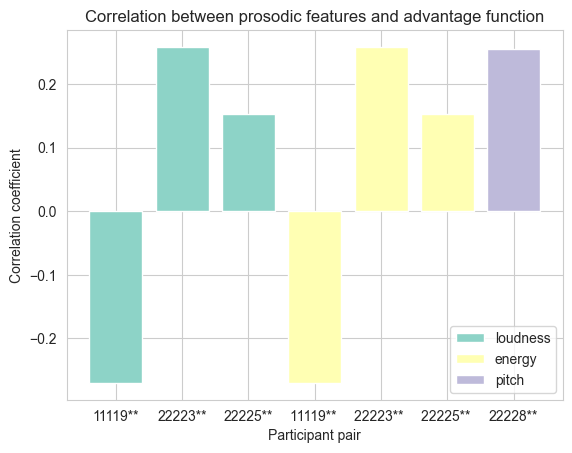}
    \caption{Statistically significant correlations between the prosodic features and the advantage values}
    \label{fig:corr_adv}
   \vspace{-9mm}
\end{figure}

The correlation between pitch and advantage values was significant for one out of fifteen participants, with a correlation coefficient of $0.25$. As mentioned above, this suggests that people give positive feedback with a higher pitch if the taken action was the best possible one.
Additionally, we conducted a repeated measure correlation. The result showed that the correlation did not hold at the population level ($p = 0.51$). Figure \ref{fig:corr_adv} shows statistically significant correlations.

The correlations for repetition and duration were both not significant. Thus, neither of the two prosodic features was associated with the advantage values.

\subsubsection{Effect of demographics}
Additionally, we investigated whether training experience can be associated with being a good agent trainer. We took the item "To what extent does your profession involve teaching or training other people?" and correlated the Likert-scale answers ranging from $1$ (None at all) to $5$ (a great deal) with the performance metric. The correlation was statistically significant ($r=-0.68$, $p = 0.02$). Thus, having a profession that involves teaching can be associated with being a good trainer. We did not find any statistically significant associations between training competence and other demographic variables such as having a pet.

\subsection{Algorithmic Implications}

Building on our empirical results, we tested the applicability of a prosody-sensitive learning paradigm by incorporating the combined speech and prosody feedback into a human reward function of an intRL agent. Since previous studies have shown promising results with intRL agents trained by using voice feedback \cite{cruz-2015} and prosodic feedback \cite{kim}, we built on this research by demonstrating that added prosody offers additional benefits beyond explicit voice feedback.

We incorporated prosody in an interactive RL algorithm and tested it in an offline manner on a subset of the sessions collected in our experiment (only the four sessions were no interpolation was needed). The intRL algorithm we chose was TAMER \cite{knox-2009} due to its popularity, and compatibility with many different feedback modalities \cite{knox-2009,empathic-framework,coach}. For implementation details, see Appendix~\ref{appendix-tamer}.

Our implemented intRL algorithm took combined explicit and implicit voice input as human feedback to learn the feedback function \textit{H}. We implemented a prosody-augmented version of TAMER while 
accounting for individual variations by normalizing it with respect to each individual's baseline mean and standard deviation (z-standardization). 
The feedback function was determined by taking the mean of all three prosodic feature values for both positive and negative feedback. We decided to make one combined prosodic feature value based on the finding that the features are highly correlated, which implies that they are used together.

The performance of the algorithm was evaluated by assessing the learned policy of \textit{H}. For this, all possible optimal actions for each state in the environment were determined. The policy for \textit{H} was determined by choosing each action using a greedy approach. The absolute number of optimal actions was counted and used as the evaluation metric.

\begin{figure}
    \centering
    \includegraphics[width = 0.3\textwidth]{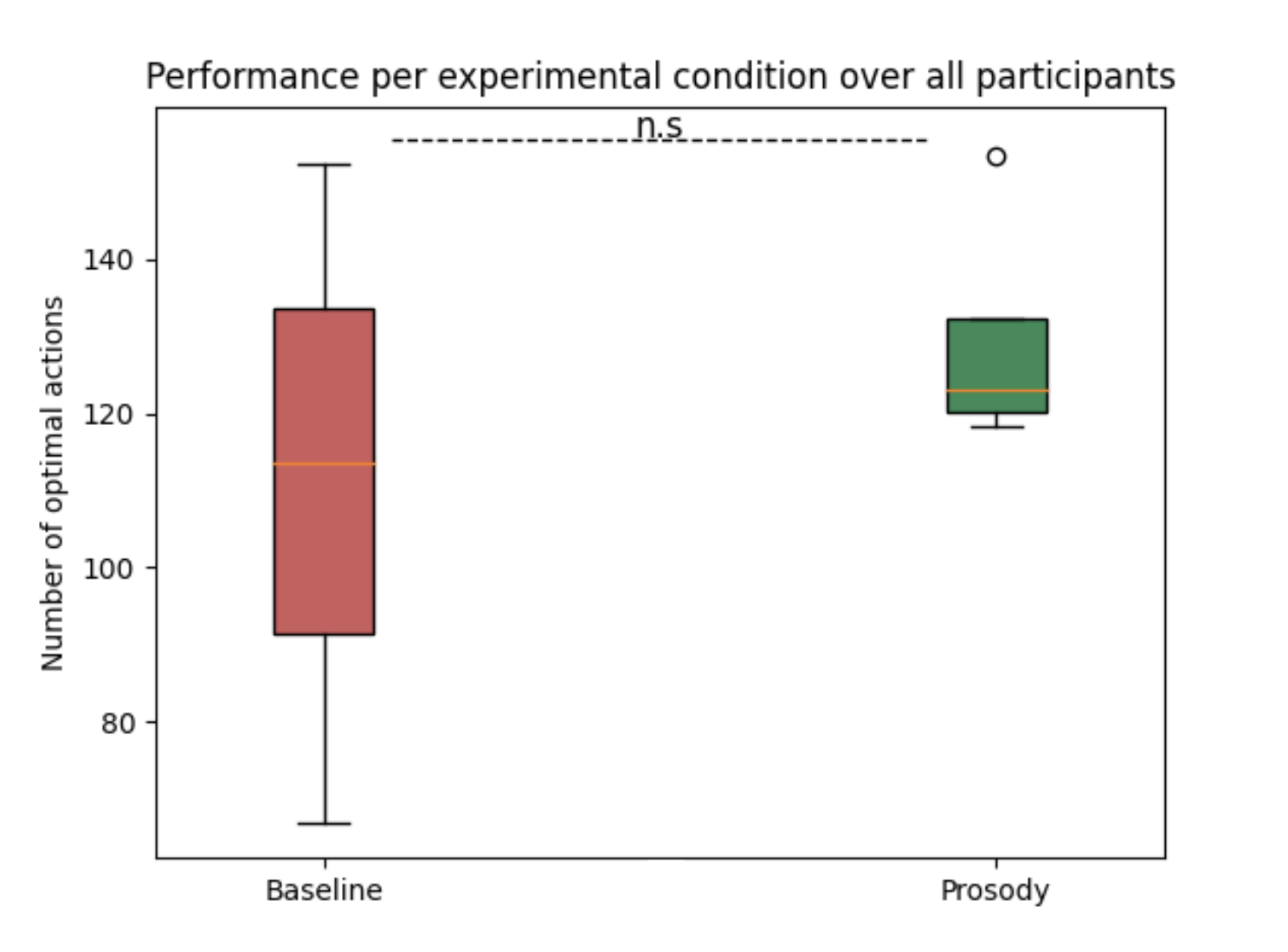}
    \vspace{-2mm}
    \caption{Performance of prosody-augmented TAMER (i.e., speech content + prosody) versus TAMER baseline (i.e., only speech content) in an offline RL setting ($N=4$)).}
    \label{fig:results_model}
    \vspace{-3mm}
\end{figure}

In Fig.~\ref{fig:results_model}, the performances are visually displayed across the four participants. At the population level, although not statistically significant due to small sample size, the prosody condition had the highest mean value of optimal actions. However, when looking at individual participants, for three out of the four participants, the prosody condition statistically significantly outperformed the TAMER baseline. These preliminary results are promising and will be followed up on in larger studies in the future.

\section{Study 2: Prosody in Audio-Augmented Demonstrations}
\label{demo-study}

In this study, we collected audio data from a single demonstrator playing three Atari games while providing only ``yes'' and ``no'' utterances. The demonstrator's prosodic features were analyzed alongside game dynamics to understand the relationship between prosody and task performance. Particularly, by restricting the usage of words (``yes'' and ``no'' only) and working with a single demonstrator, we isolate the impact of speech prosody from spoken words and user differences respectively.

\subsection{Study Design}
\subsubsection{Experimental Setup}
\begin{figure}
    \centering
    \includegraphics[width=0.5\textwidth]{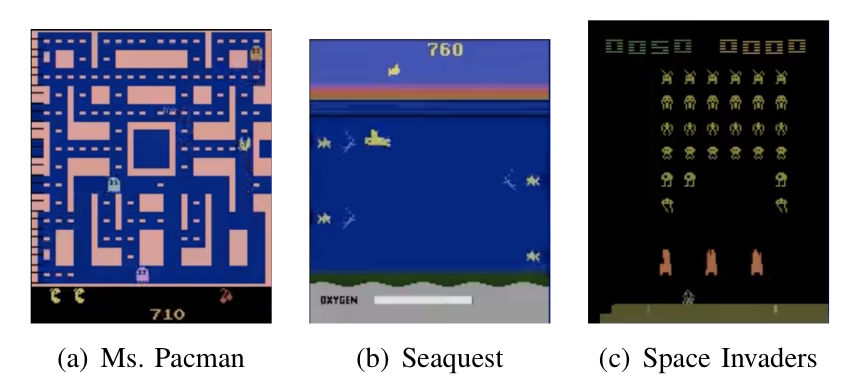}
    \vspace{-7mm}
    \caption{Atari games used in our study to understand a human demonstrator's speech patterns.}
    \vspace{-3mm}
\label{fig:audio_games}
\end{figure}

We collect demonstration and audio data from a single demonstrator using a customized simulation interface for three Atari Games shown in Figure~\ref{fig:audio_games}: (1) Ms. Pac-Man, (2) Seaquest, (3) Space Invaders. We use these games due to the diversity in their objectives and reward schemes. The demonstrator provides state-action data (states are images of the game screen, and actions are keystrokes used for game-play) and audio data in the form of only ``yes'' and ``no'' utterances (collected over the web via the demonstrator’s device microphone) for 10 minutes of gameplay per game. This data was collected remotely through a web interface and recorded on a remote server. Screenshots of the data collection interface are shown in Appendix~\ref{appendix:screenshots_atari}. We asked the user to demonstrate examples of how to play the games to the best of their ability by using their keyboard as well as by using their microphone so the Atari agent will learn how to play the game by observing both their keyboard strokes as well as their voice. We also stated that the character will move and be controlled by their keystrokes only, but the agent can understand the words ``yes'' and ``no'', and can identify the pitch or intensity with which they say these words. 

\subsubsection{Procedure}
The user was provided an opportunity to practice the game before their demonstrations were recorded to help them better understand the dynamics and reward structure of the game. The user was instructed to play the games in a quiet environment to minimize the possibility of any background noise being recorded. Data is collected using the Atari Grand Challenge (AGC) interface~\citep{kurin2017atari}. However, an additional functionality is added to record the human audio synchronized with demonstrated states and actions (Appendix~\ref{appendix:screenshots_atari}). The start and end times of each utterance are detected with Google’s speech-to-text API
, and the reward values are provided by the logs of the AGC interface. 

\subsection{Measures}
\textit{\textbf{Prosodic Features:}} After collecting and synchronizing the state, action, and speech data, prosodic feature values for each utterance were computed by loading the data in the python library Librosa and a default sampling rate of 22050 \cite{librosa}. We use \textit{utterance duration}, \textit{utterance repetition}, \textit{energy}, \textit{pitch}, and \textit{loudness} as the different prosodic features based on prior work in the speech recognition and learning from demonstration communities~\citep{hirschberg2004prosodic, saran2022understanding}.  
Following these, we report the mean and maximum values for pitch, loudness, and energy for each speech utterance. We also report the total energy (cumulative energy sum per utterance).

\textit{\textbf{MDP Features: }}In prior work studying unrestricted speech prosody for robot learning by \citet{saran2022understanding}, demonstration rewards or errors are shown to be the most promising MDP statistic to leverage with prosody to enhance learning. Thus, we study how often the demonstrator uses ``yes'' and ``no'' utterances, as well as the connection of prosody with the underlying ground truth rewards in the game.

Based on prior studies that have identified that prosodic features can convey semantic meaning for words used in speech~\citep{nygaard2009semantics}, we expect that the prosodic feature values can help identify the nature of speech used (positive/negative) by being significantly different for the ``yes'' and ``no'' utterances. The meaning or nature of speech used is also evaluated against the underlying ground truth rewards from the game design.

\subsection{Results}
\label{sec:atari_study_results}
\subsubsection{Positive/Negative bias:} 
From the playback of synchronized audio along with demonstrations, we observe that the demonstrator uses utterances as both a reaction to recent past events and anticipatory guidance for future events during a demonstration. For this reason, we use a $0.5$ second buffer before the start time and after the end time of each audio utterance when computing features or reward values accompanying an utterance.  The average duration of utterances is $0.76$ seconds (Ms. Pacman), $0.61$ seconds (Seaquest), and $0.75$ seconds (Space Invaders) (Fig.~\ref{fig:utt_dur_atari}). 

We compute the overall duration of ``yes'' and ``no'' utterances compared to the total duration of a demonstration (see Fig.~\ref{fig:yes_no_percentage} in Appendix~\ref{appendix:atari_results}). We find that the demonstrator uses significantly more ``yes'' utterances compared to ``no'' utterances (Fig.~\ref{fig:yes_no_percentage}) for all three games ($p<0.05$). 
The proportion of ``yes'' utterances is the highest for Ms. Pacman which is a multi-objective game where the agent has to move rapidly by escaping ghosts and procuring food palettes. There is less room for respite when the game begins, which might have made the demonstrator more active with their speech in this game compared to other games. Thus, ``no'' utterances or negative feedback are used for rare events during the games, where the agent loses points or dies (end of episode), whereas ``yes'' or positive feedback is used more frequently to indicate good progression in the game.

Next, we compute prosodic feature values accompanying the ``yes'' and ``no'' utterances for each game (see Fig.~\ref{fig:prosody_yes_no} in Appendix~\ref{appendix:atari_results}).
We find that mean energy, maximum energy, total energy, mean loudness, and maximum loudness are significantly higher ($p<0.05$) for negative feedback (``no'') compared to positive feedback (``yes'') for all the games (except maximum loudness during Space Invaders).
This finding is similar to that of Sec.~\ref{sec:results_intrl_pos_neg} for interactive RL.

\subsubsection{Link to MDP statistics:}
The cumulative sum of rewards for trajectory snippets accompanying speech utterances is significantly higher for ``yes'' utterances versus ``no'' utterances (Fig.~\ref{fig:yes_no_return}). This finding intuitively states that the cumulative reward during positive feedback (``yes'') is higher than the cumulative reward during negative feedback (``no'') from the demonstrator.

\begin{table}[!htb]
    \centering
    \vspace{-2mm}
    \caption{Spearman correlation between mean pitch and ground truth rewards for ``yes'' and ``no'' utterances used along with demonstrations to three different Atari games. 
    }
    \vspace{-2mm}
    \begin{tabular}{|c|c|c|}
    \hline
    & Yes & No\\
    \hline
        \hline
        Ms. Pacman & $0.13$ & $-0.12$ \\
        \hline
        Seaquest & $0.2^{*}$ & $-0.77^{*}$\\
        \hline
        Space Invaders & $0.37^{**}$ & $-0.38^{*}$\\
        \hline
    \end{tabular}
    \vspace{-2mm}
    \label{tab:correlation_prosody}
\end{table}

While mean pitch is not significantly different for ``yes'' and ``no'' utterances (Fig.~\ref{fig:prosody_yes_no}), we find another pattern for mean pitch (but not for other prosodic features)--- there is a positive spearman correlation between mean pitch and cumulative sum of rewards for trajectory snippets accompanying ``yes'' audio utterances (Table~\ref{tab:correlation_prosody}). Similarly, we also observe a negative spearman correlation between mean pitch and cumulative rewards for trajectory snippets accompanying ``no'' audio utterances. This finding reveals that ``yes'' utterances with higher pitch have higher ground truth returns associated with them and ``no'' utterances with higher pitch have lower returns associated with them, i.e. the more the gain or loss with an action during the demonstration, the more emphatically the corresponding word is spoken. We do not find a consistent pattern of correlation for any other prosodic features such as energy, loudness.
Thus, while pitch might not distinguish between the yes and no utterances, it can correlate with the magnitude of reward values based on the type of speech usage (positive/negative). These results hint towards prosody in speech revealing information about the underlying rewards, thereby potentially enhancing the sample efficiency for LfD methods which first learn the underlying reward function to train the agent policy (inverse reinforcement learning).

\subsection{Algorithmic Implications}

While several recent works \citep{zhang2019cutting, liang2019making, gandhi2019swoosh} have utilized object or environmental audio for learning from demonstration, to the best of our knowledge, spoken and prosodic cues of human teachers have not been leveraged with deep LfD algorithms. Based on the findings in Sec.~\ref{sec:atari_study_results}, we propose an efficient technique to leverage speech in the form of an auxiliary loss for reward learning. The auxiliary loss leverages both spoken words and prosodic features from human speech to guide the training of a deep inverse reinforcement learning method T-REX~\citep{brown2019extrapolating} for the Atari game-playing domain.
T-REX trains a deep reward model by comparing the performance of pairs of demonstrated trajectory snippets. The final agent policy is trained via an RL algorithm (such as PPO~\cite{schulman2017proximal}) on the learned reward.
Since the only two words used by the demonstrator signify if an event during the demonstration is positive (yes) or negative (no) as shown in Sec.~\ref{sec:atari_study_results} and Fig.~\ref{fig:yes_no_return} in Appendix~\ref{appendix:atari_results}, the ``yes'' and ``no'' labels can create contrasting categories of sample trajectory snippets for reward learning. The reward values predicted by the reward network can be compared for trajectory snippets of a similar spoken word, and in turn the reward network's parameters can be appropriately penalized during training. We use the prosodic feature values to determine how to scale the similarity between pairs of snippets accompanying audio utterances. Based on the finding in Sec.~\ref{sec:atari_study_results} and Table~\ref{tab:correlation_prosody}, correlation of mean pitch with reward values motivate us to scale up the similarity between the rewards for two demonstration snippets if the difference in their corresponding pitch is small.

\subsubsection{Details of Model Training}
The audio, state, and action data are synchronized post data collection to accurately extract demonstrated trajectory snippets accompanied by audio utterances for training the T-REX agent~\citep{brown2019extrapolating}.
We use a contrastive loss to guide the training of the reward network for T-REX~\citep{brown2019extrapolating} which we call the contrastive audio loss (CAL).  Given a sequence of $K$ demonstrations ranked from worst to best,  $\tau_1, \dots ,\tau_K$, a parameterized reward network $\hat{r}_\theta$ is trained with a cross-entropy loss over a pair of trajectories ($\tau_i \prec \tau_j$), where $\tau_j$ is ranked higher than $\tau_i$. We add CAL as an auxiliary loss for training the reward network with additional trajectory pairs $\tau_m$ and $\tau_n$ which are accompanied by the same audio utterances, so the new loss function becomes:
\vspace{-1mm}
\begin{equation} 
\begin{split}
&\mathcal{L}(\theta) = - \sum_{\tau_i \prec \tau_j} \log \frac{\exp \sum_{s \in \tau_j}\hat{r}_\theta(s)}{\exp \sum_{s \in \tau_i}\hat{r}_\theta(s) + \exp \sum_{s \in \tau_j}\hat{r}_\theta(s)} \hspace{3mm} +  \\ & \alpha \Bigg [  CAL\Big(\sum_{s \in \tau_m}\hat{r}_{\theta}(s),p_{m}, \sum_{s \in \tau_n}\hat{r}_{\theta}(s),p_{n}\Big) \Bigg ] 
\end{split}
\end{equation}

where
\begin{equation}
    CAL\Big(R^{a}_{m},p_{m},R^{a}_{n},p_{n}\Big) = -\log\frac{\exp(\text{sim}(R^{a}_{m}, R^{a}_{n})/t_{mn})}{\sum^{N}_{k=1}\exp(\text{sim}(R^{a}_{k_0}, R^{a}_{k_1})/t_{k_0k_1})} 
\end{equation}
\begin{equation}
    t_{mn} = t_0 + |p_m-p_n|
\end{equation}
\begin{equation}
    \text{sim}(R^{a}_{m}, R^{a}_{n}) = \frac{1}{1+|R^{a}_{m}-R^{a}_{n}|}
\end{equation}

$\sum_{s \in \tau_m}\hat{r}_{\theta}(s)$ (Equation 1) or $R^{a}_{m}$ (Equation 2) represent the undiscounted cumulative sum of (estimated) rewards corresponding to every state $s$ from the trajectory snippet $\tau_m$ accompanied by audio $a$. $\text{sim}(R^{a}_{m}, R^{a}_{n})$ in Equations 2,4 represents a similarity measure between the undiscounted cumulative sum of rewards for two trajectory snippets, and $p_{m}$ in Equations 2,3 represents a scalar prosodic feature value for the audio chunk $a_m$ accompanying $\tau_m$. The difference in prosodic features $|p_m-p_n|$ is normalized with a softmax function for all the differences from a batch. $t_{mn}$ in Equations 2,3 is the temperature parameter, and $t_0$ is an offset temperature value to avoid numerical inconsistencies. There are a total of N audio snippet pairs of dissimilar spoken words (yes/no) to compare against in the denominator of Equation 2. Snippets from each pair are denoted as $k_0, k_1$ where $k\in[1,N]$. Based on findings in Sec.~\ref{sec:atari_study_results}, we use mean pitch as the prosodic feature to determine the temperature values automatically. The closer two audio chunks are (for a specific spoken word, i.e. either only yes or only no) in terms of their corresponding pitch values, the closer their reward values should be. 
The loss function accumulates the impact of audio over the entire trajectory snippet for each trajectory pair used as input to the network.

We evaluate the performance of each game with and without CAL augmentation during training. Performance is measured in terms of game score averaged over $30$ trials for three seeds, with the highest result reported among the three seeds. Standard error is also reported along with mean performance.

\subsubsection{Implications}
Results for the three games are shown in Table~\ref{tab:cal_results}, where incorporating speech cues via CAL improves average performance of each game. This highlights the efficacy of incorporating speech cues from a single demonstrator for learning from demonstration. 
Both the content of speech (what is being said), and prosody (how something is said) are useful in guiding agents trained via the T-REX algorithm. 
Here, the content of words have a direct mapping to positive and negative events when comparing a pair of trajectory snippets, whereas prosody is used in a manner that scales the similarity between two spoken utterances (and in turn the similarity in the predicted reward values) based on the magnitude of their corresponding pitch.  

\begin{table}
    \centering
    \caption{Average performance over 30 rollouts for three Atari game agents trained in the LfD paradigm.
    }
    \vspace{-2mm}
    \begin{tabular}{|c|c|c|}
    \hline
    & T-REX & T-REX + CAL\\
    \hline
        \hline
        Ms. Pacman & 414.0$\pm$14.9 & \textbf{663.3$\pm$121.3} \\
        \hline
        Seaquest & 679.3$\pm$14.3 & \textbf{704.0$\pm$6.0}\\
        \hline
        Space Invaders & 1235.2$\pm$102.3 & \textbf{1781.3$\pm$127.6} \\
        \hline
    \end{tabular}
    \vspace{-5mm}
    \label{tab:cal_results}
\end{table}

In summary, we find that the content of speech is more indicative of the underlying reward type (low/high), while prosodic cues are more indicative of scale of rewards. We take the findings from this study and build on them by augmenting a deep imitation learning algorithm with speech.
We present an auxiliary loss (contrastive audio loss) to leverage simple predefined speech cues from a single demonstrator to train T-REX agents for high-dimensional Atari games. Our investigation shows that underutilized speech cues can effectively guide agent learning in the learning from demonstration paradigm. Leveraging the underutilized speech modality can thus enable sample efficient training and save time spent by human teachers demonstrating tasks to AI agents (extracting more information from a fixed set of demonstrations with minimal cost to record speech information and without requiring any additional time spent by the teacher).

\section{Discussion}
Speech is a low-effort and rich source of information that humans naturally provide when teaching artificial learning agents. In addition to spoken words from natural language, speech also contains prosodic cues which can be informative towards demystifying the underlying goals or progress during the task to artificial learning agents. Our exploratory studies about the integration of prosody as a teaching signal in agent learning environments have yielded promising results, illustrating the potential of prosodic cues to enhance the efficacy of human-agent interactions in both reinforcement learning and learning from demonstration scenarios. 

One of the primary limitations of our studies relates to the controlled nature of the feedback environment, where we restricted the verbal input to binary ``Yes''/``No'' responses. While this design choice was intended to minimize confounding variables, it also limits the richness of the feedback and may not fully represent more dynamic real-world interactions where verbal feedback can be more nuanced. The generalizability of our results across different tasks and agent embodiments (beyond simulations) also remains an open question. Our studies were conducted with a specific set of tasks, and future work should explore how well our findings would translate to other contexts or more complex decision-making environments.

To better account for variations across human teachers, future studies could incorporate adaptive learning systems that are sensitive to individual teaching styles and learner responses. Such systems can adjust their interpretation of prosodic cues based on the specific teacher-learner dyad, potentially through personalized calibration sessions or real-time feedback mechanisms. Future research could also explore the integration of richer verbal feedback and the development of more sophisticated models for prosody recognition and interpretation. The potential for real-time adaptation to the teacher's prosodic patterns offers an intriguing avenue for developing more responsive and sensitive learning agents. Moreover, incorporating multi-modal feedback, where prosodic cues are considered in conjunction with other non-verbal cues such as gaze, gestures, or facial expressions, could provide a more holistic approach to understanding and leveraging natural human communication patterns for interactive machine learning.

\section{Conclusion}
In this work, we present analyses of human prosody during two interactive learning paradigms --- interactive reinforcement learning (intRL) and learning from demonstration (LfD). We find that prosody from human teachers is expressed more strongly during negative feedback compared to positive feedback in both learning paradigms. Additionally, we find correlations between prosodic feature values and the advantage function for intRL, and between prosodic feature values and the reward function for LfD. With these results, we motivate the design of novel algorithms in the two learning paradigms which leverage human prosody during learning. Our proof of concept experiments reveal that prosody is a promising modality to enhance learning and improve sample efficiency.

This paper highlights the untapped potential of prosody in enhancing agent learning from human interaction. By advocating for the integration of prosody-sensitive algorithms and providing empirical evidence of their efficacy, we aim to advance the field of human-agent interaction and pave the way for more intuitive and efficient learning paradigms.
\begin{acks}
The authors would like to thank Taylor Kessler Faulkner, Andrea Thomaz, Scott Niekum for their valuable feedback, and Ojas Patel, Rakesh Johny for contributing to the codebase.
\end{acks}

\bibliographystyle{ACM-Reference-Format}
\bibliography{sample-base}

\clearpage

\appendix

\section{Appendix: Interactive RL}
\label{appendix-interactive-rl}

\subsection{Data Collection Interface (Teacher)}
\label{appendix-teacher}

Below we share screenshots of the data collection interface provided to the teacher during a user study session.

\begin{figure}[h]
    \centering
    \begin{subfigure}
        \centering
        \includegraphics[width=0.95\textwidth]{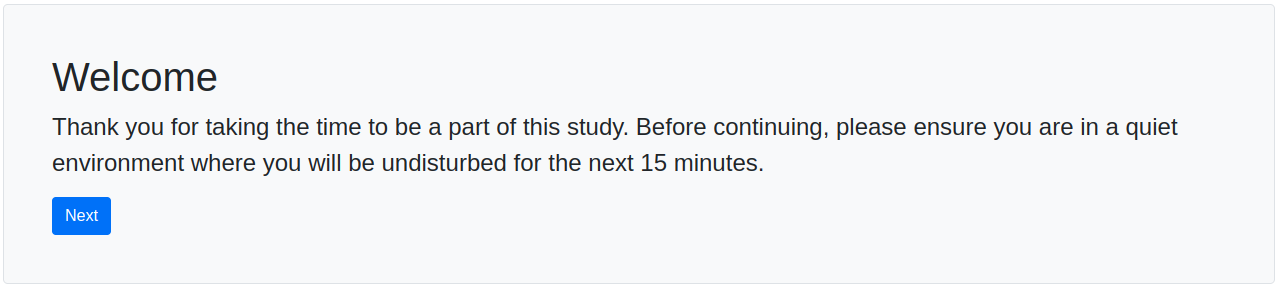}
        \label{fig:intrl_teacher_1}
    \end{subfigure}%
    ~ 
    \begin{subfigure}
        \centering
        \includegraphics[width=0.95\textwidth]{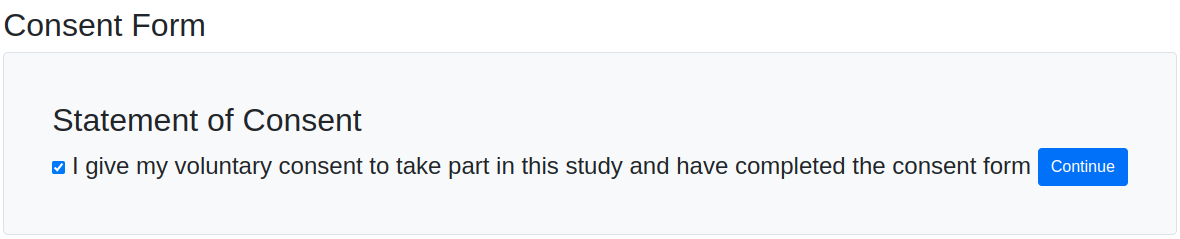}
        \label{fig:intrl_teacher_2}
    \end{subfigure}%
    ~
    \begin{subfigure}
        \centering
        \includegraphics[width=0.7\textwidth]{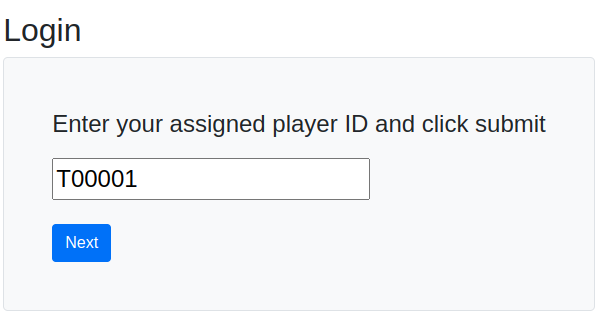}
        \label{fig:intrl_teacher_3}
    \end{subfigure}
\end{figure}
\begin{figure*}[h]
    \centering
    \begin{subfigure}
        \centering
        \includegraphics[width=\textwidth]{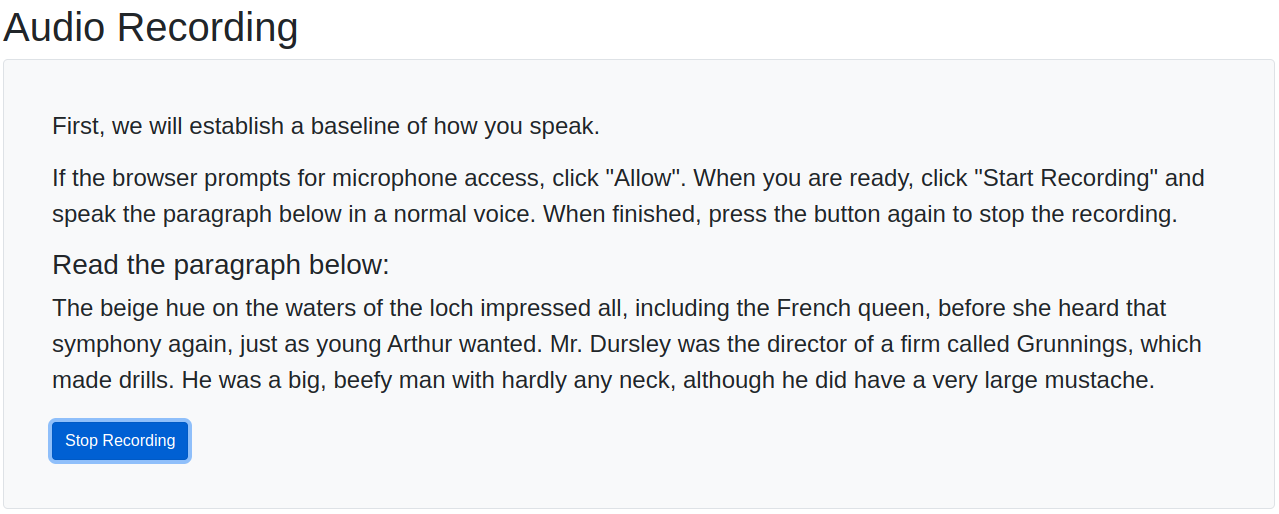}
        \label{fig:intrl_teacher_4}
    \end{subfigure}%
    ~
    \begin{subfigure}
        \centering
        \includegraphics[width=\textwidth]{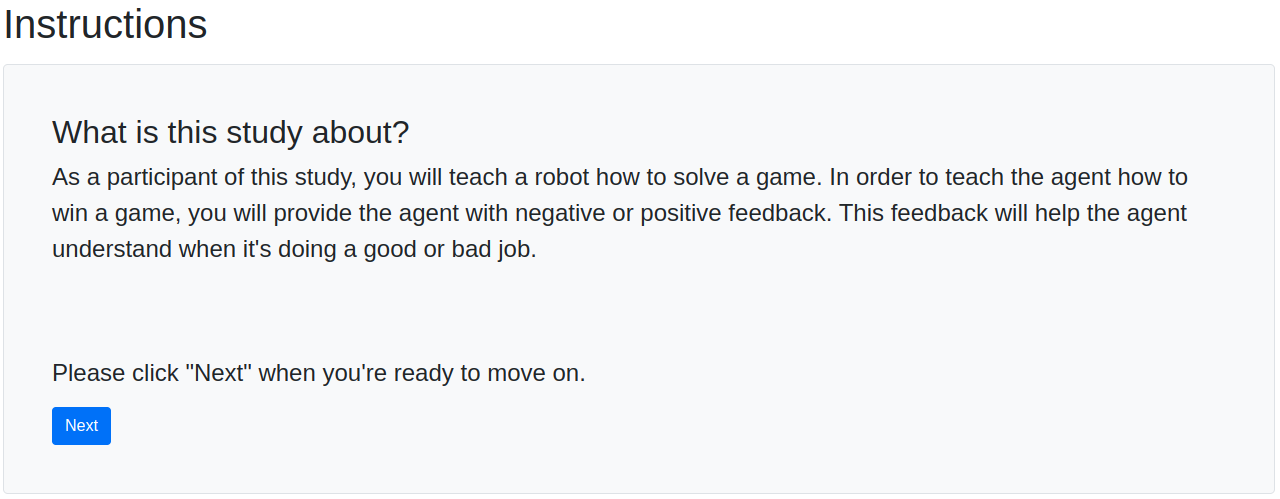}
        \label{fig:intrl_teacher_5}
    \end{subfigure}
\end{figure*}
\begin{figure*}[h]
    \centering
    \begin{subfigure}
        \centering
        \includegraphics[width=0.95\textwidth]{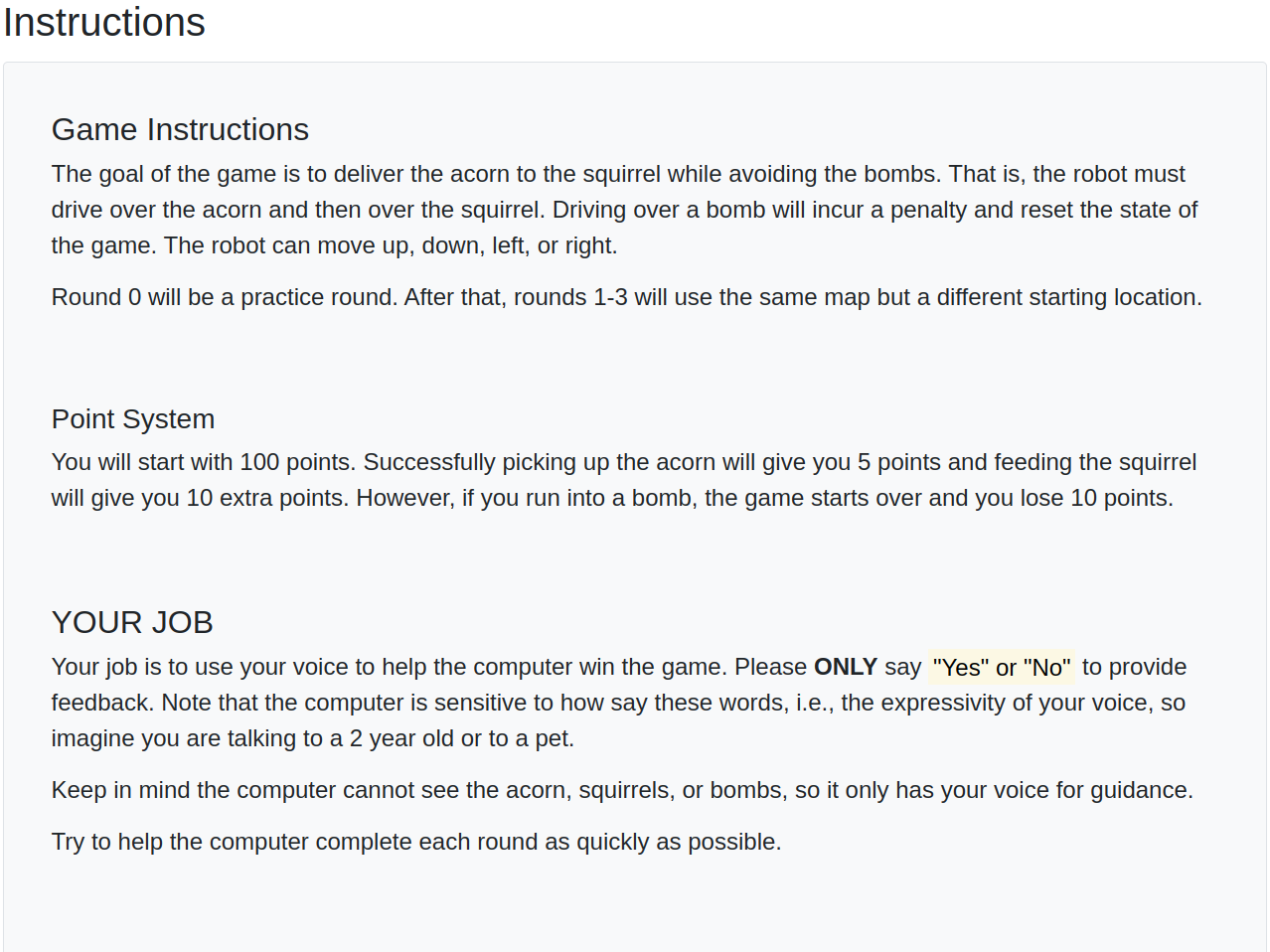}
        \label{fig:intrl_teacher_6}
    \end{subfigure}%
    ~
    \begin{subfigure}
        \centering
        \includegraphics[width=0.95\textwidth]{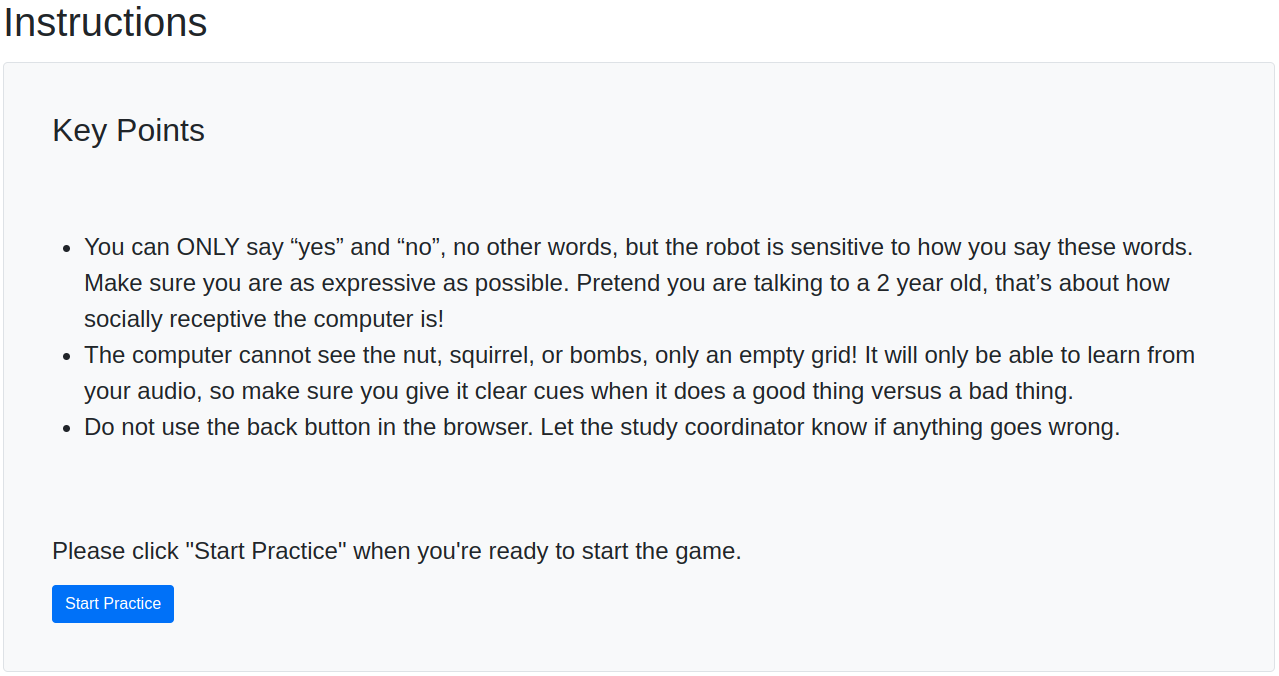}
        \label{fig:intrl_teacher_7}
    \end{subfigure}
\end{figure*}
\begin{figure*}[h]
    \centering
    \begin{subfigure}
        \centering
        \includegraphics[width=0.9\textwidth]{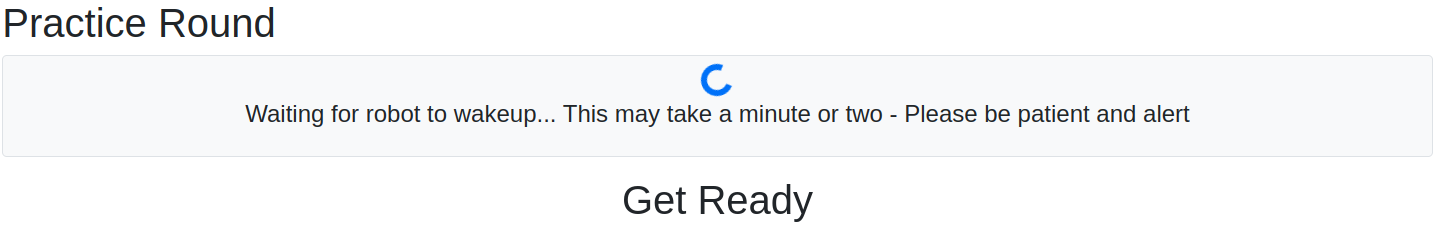}
        \label{fig:intrl_teacher_8}
    \end{subfigure}%
    ~
    \begin{subfigure}
        \centering
        \includegraphics[width=1.0\textwidth]{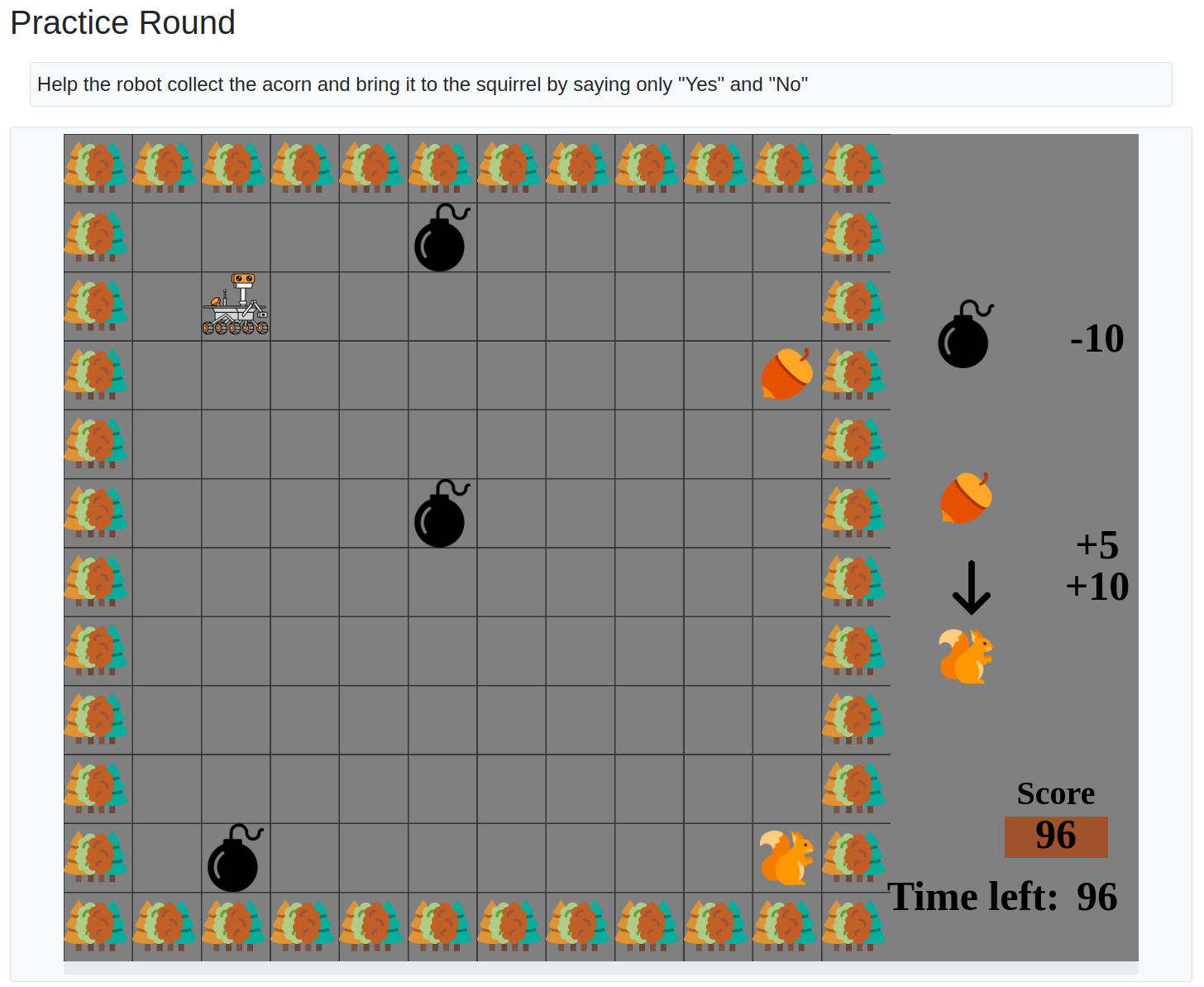}
        \label{fig:intrl_teacher_9}
        \vspace{0.2cm}
    \end{subfigure}
    \caption{The teacher is able to see all game objects and use their speech to provide feedback for both the trial and actual games. However, they are unable to control the agent}
    \vspace{0.2cm}
\end{figure*}
\begin{figure*}
    \centering
    \begin{subfigure}
        \centering
        \includegraphics[width=0.9\textwidth]{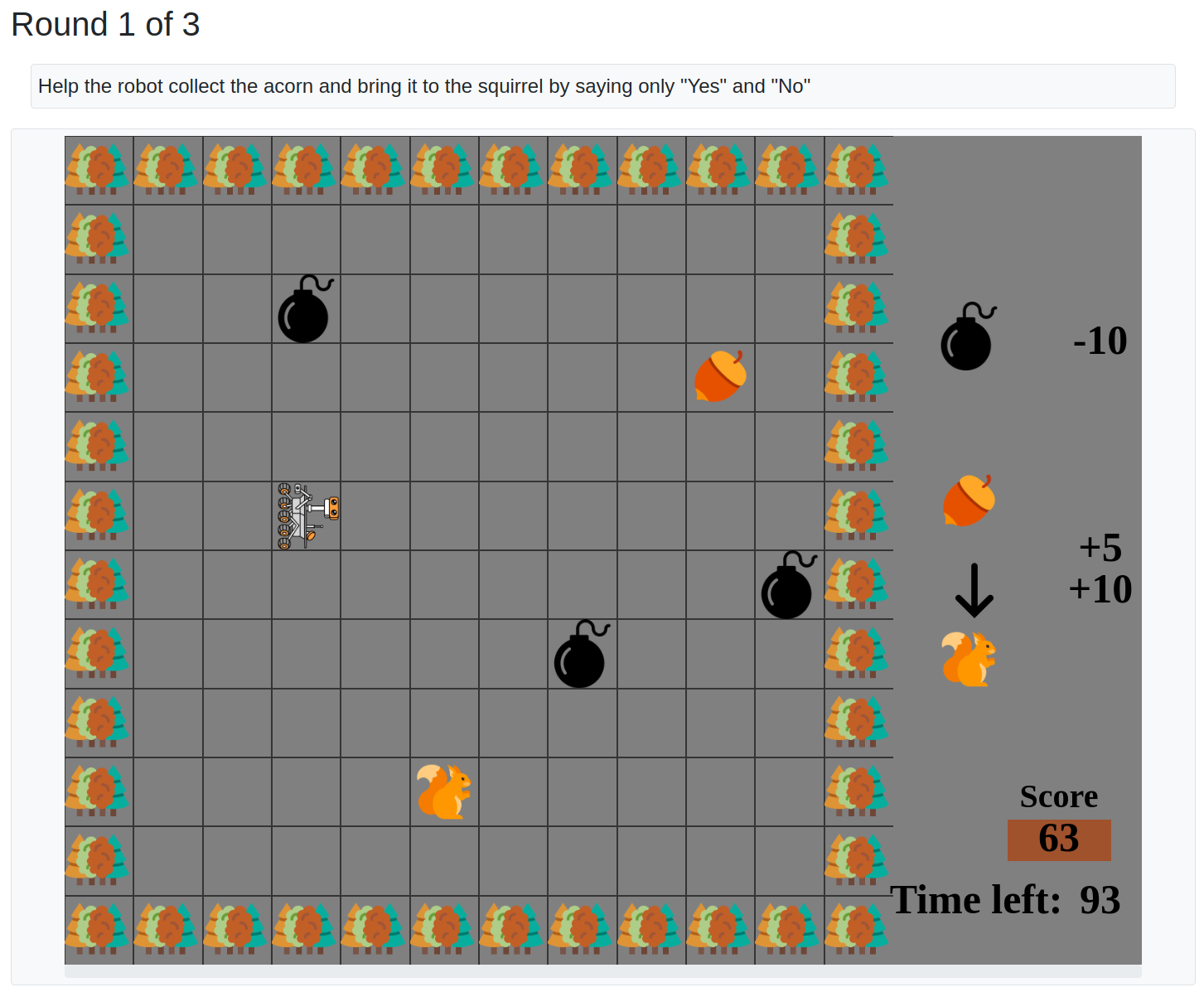}
        \label{fig:intrl_teacher_10}
        \caption{The teacher is able to see all game objects and use their speech to provide feedback for both the trial and actual games. However, they are unable to control the agent}
        \vspace{0.2cm}
    \end{subfigure}
    \begin{subfigure}
        \centering
        \includegraphics[width=0.7\textwidth]{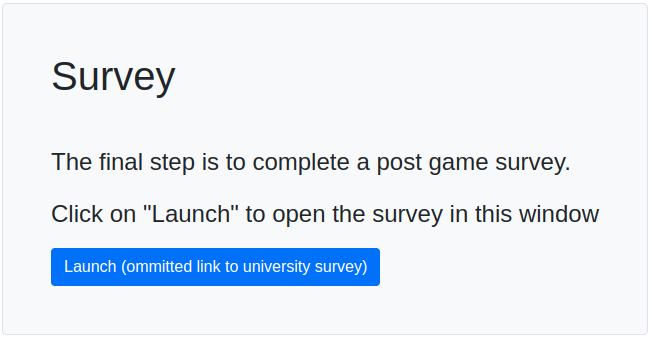}
        \label{fig:intrl_teacher_11}
        \vspace{0.2cm}
    \end{subfigure}
    \vspace{0.2cm}
\end{figure*}

\clearpage

\subsection{Data Collection interface (Wizard)}
\label{appendix-wizard}

Below we share screenshots of the data collection interface provided to the wizard during a user study session.

\begin{minipage}{\textwidth}
    \begin{figure}[H]
        \centering
        \begin{subfigure}
            \centering
            \includegraphics[width=0.95\textwidth]{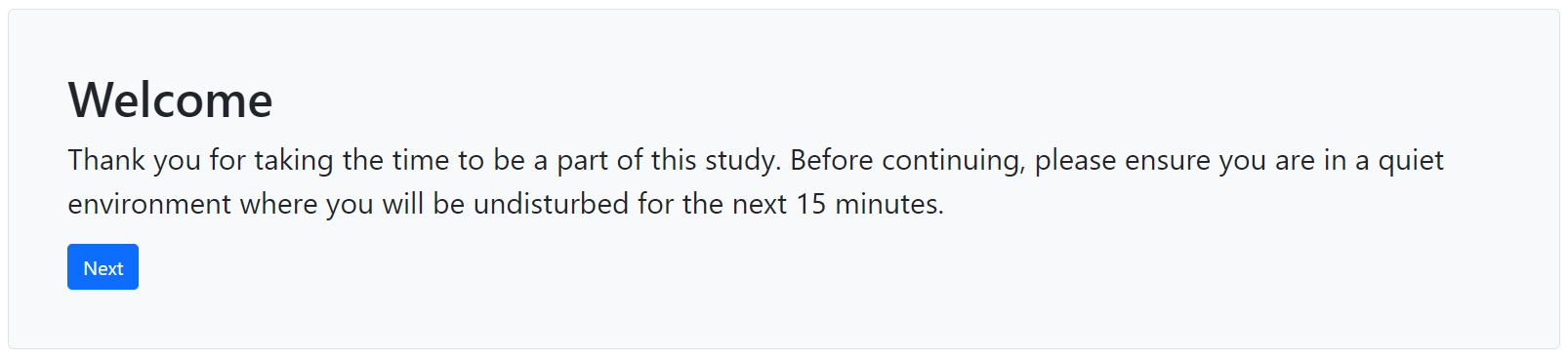}
            \label{fig:intrl_worker_1}
        \end{subfigure}
        \begin{subfigure}
            \centering
            \includegraphics[width=0.95\textwidth]{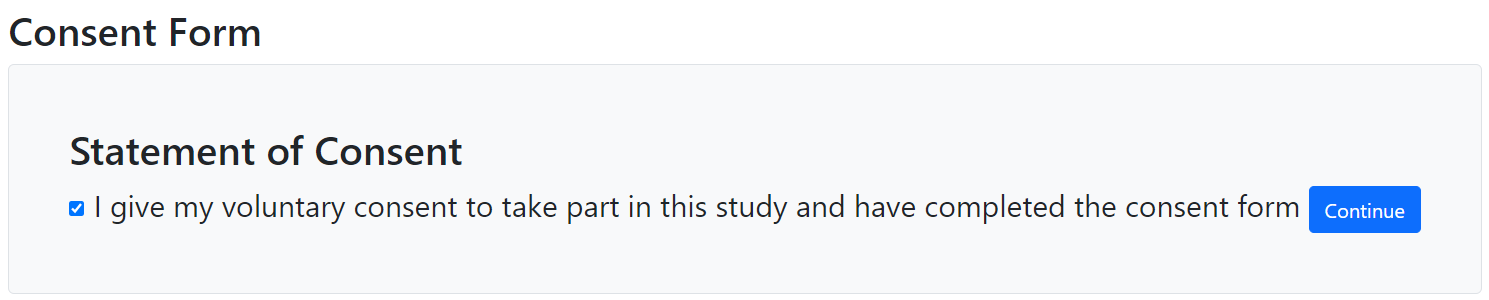}
            \label{fig:intrl_worker_2}
        \end{subfigure}
        \begin{subfigure}
          \centering
          \includegraphics[width=0.75\textwidth]{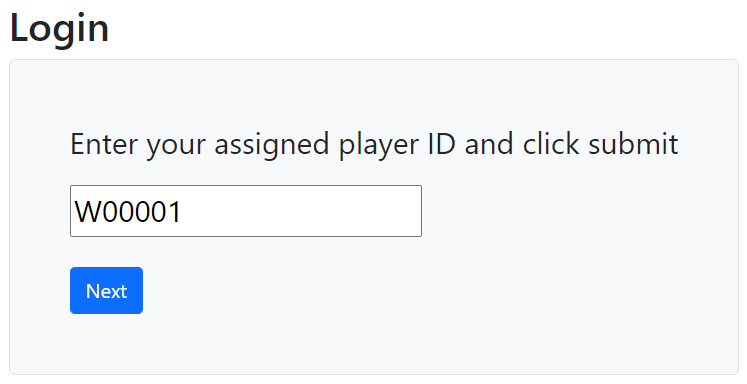}
          \label{fig:intrl_worker_3}
        \end{subfigure}
    \end{figure}
\end{minipage}

\clearpage
\begin{figure*}[h]
    \centering
    \begin{subfigure}
      \centering
      \includegraphics[width=0.9\textwidth]{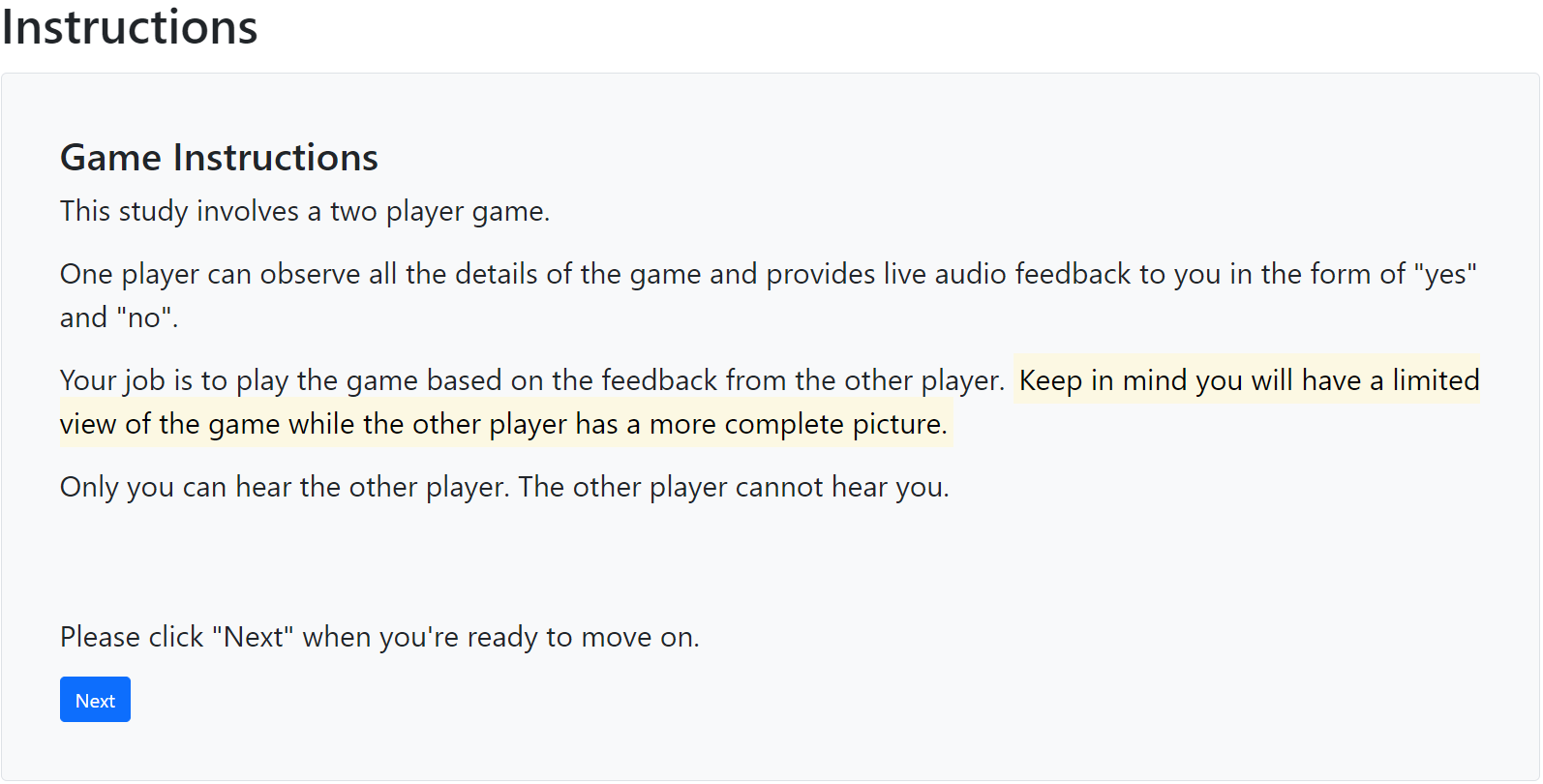}
      \label{fig:intrl_worker_4}
    \end{subfigure}
    \begin{subfigure}
        \centering
        \includegraphics[width=0.9\textwidth]{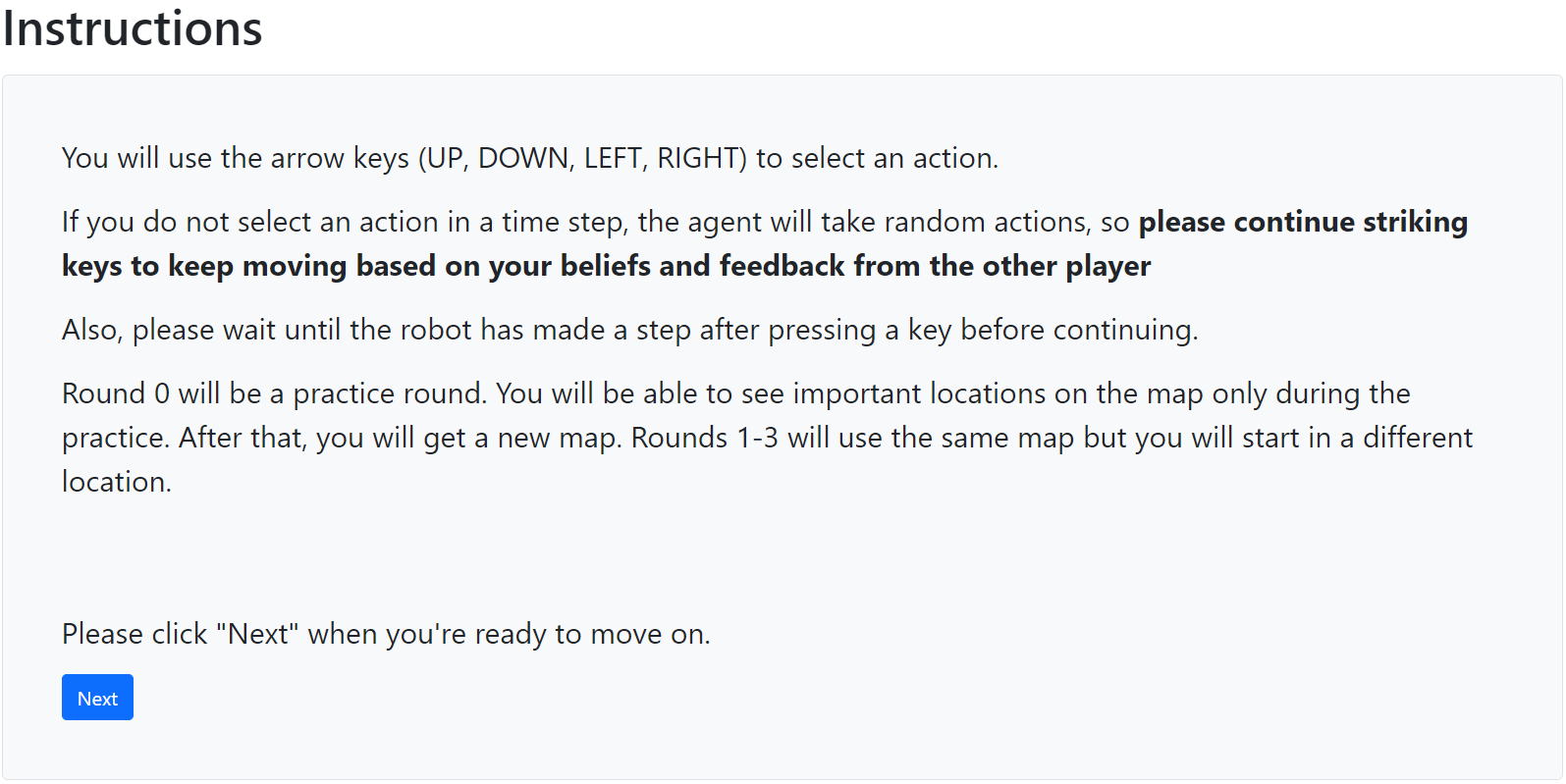}
        \label{fig:intrl_worker_5}
    \end{subfigure}
    \begin{subfigure}
        \centering
        \includegraphics[width=0.9\textwidth]{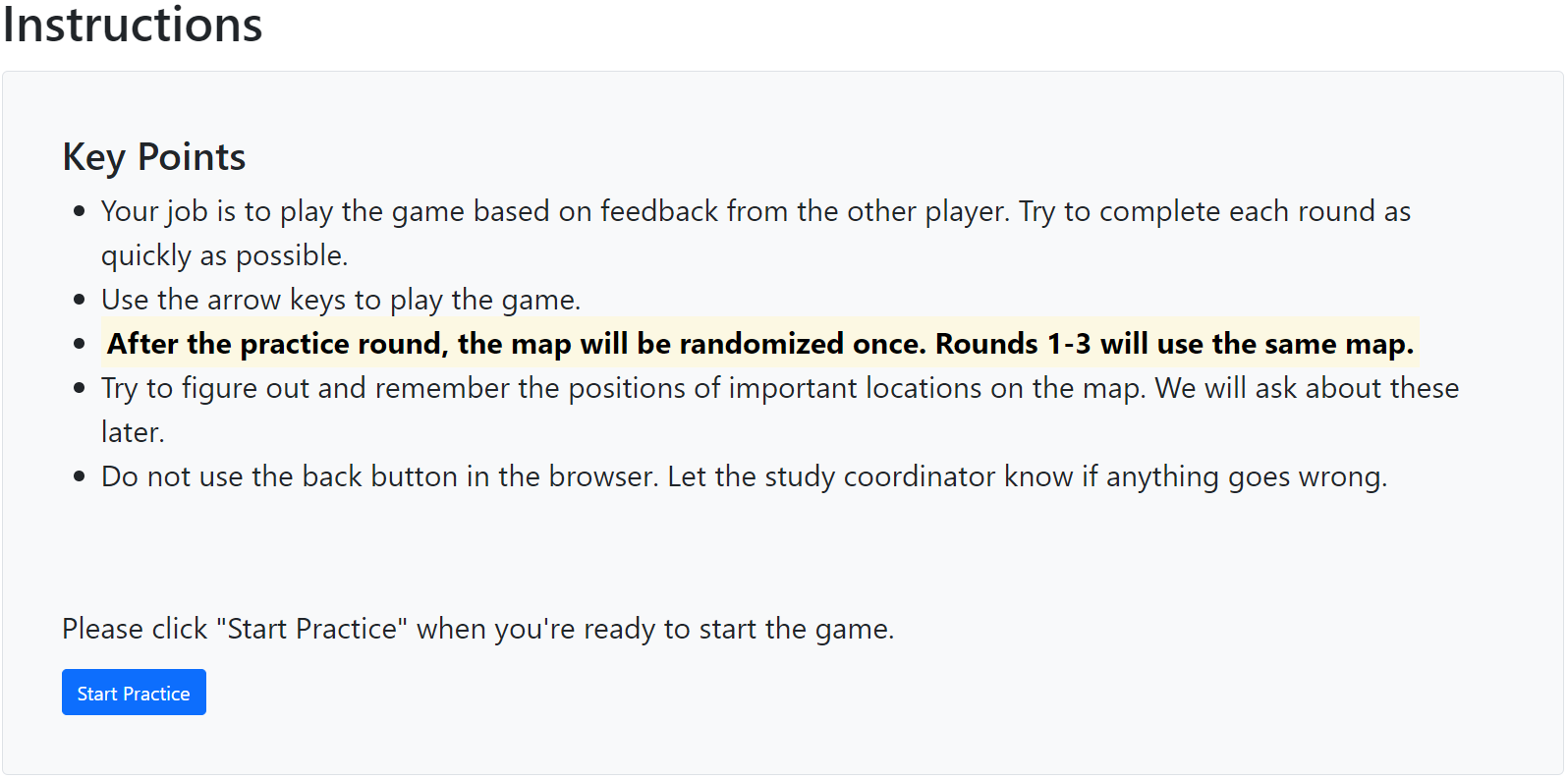}
        \label{fig:intrl_worker_6}
    \end{subfigure}
\end{figure*}
\begin{figure*}[h]
    \centering
    \begin{subfigure}
        \centering
        \includegraphics[width=\textwidth]{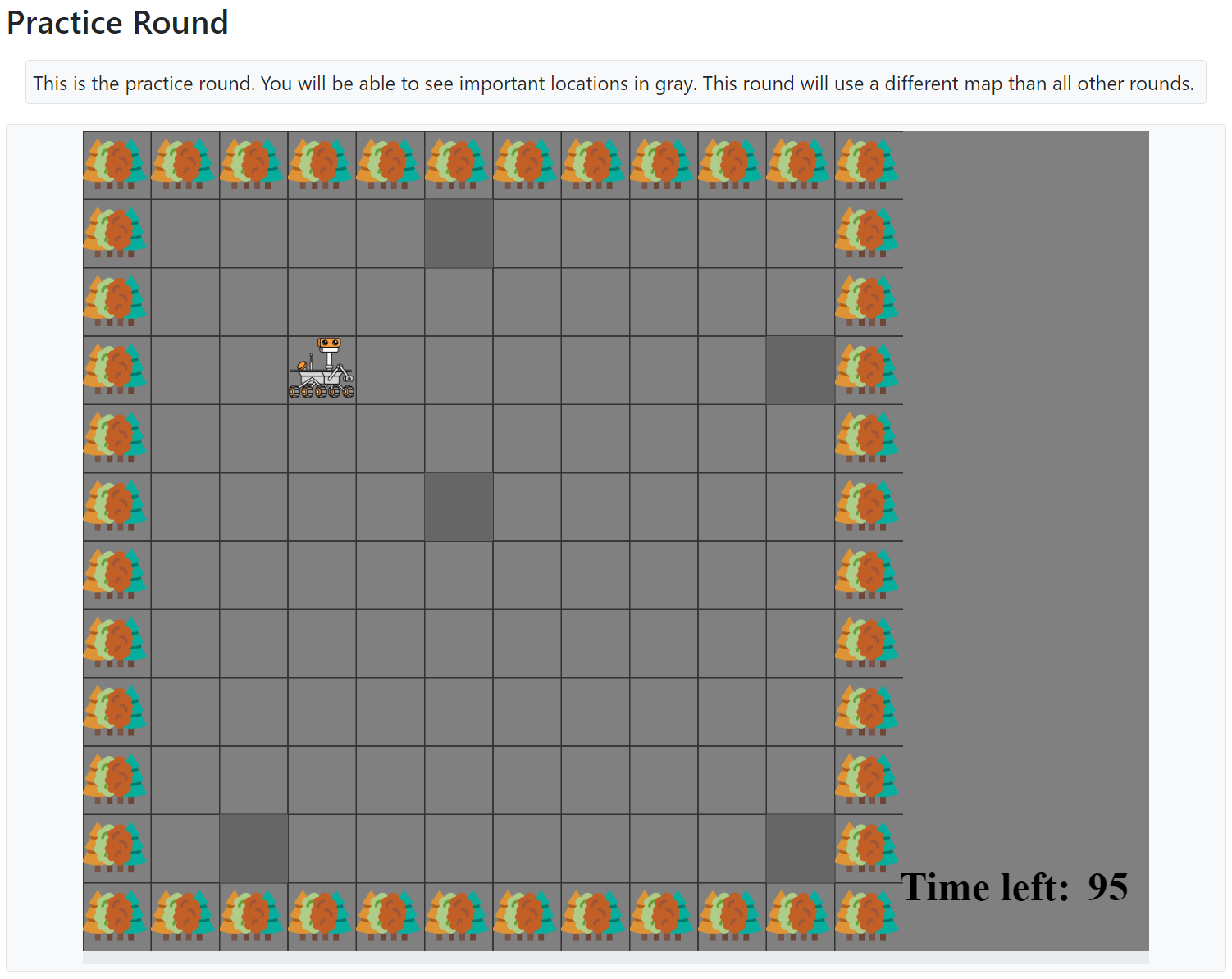}
        \label{fig:intrl_worker_8}
        \caption{In the trial period, the worker is only permitted to see the locations of the game objects but not the objects themselves}
    \end{subfigure}
\end{figure*}
\begin{figure*}[h]
    \centering
    \begin{subfigure}
        \centering
        \includegraphics[width=2\columnwidth]{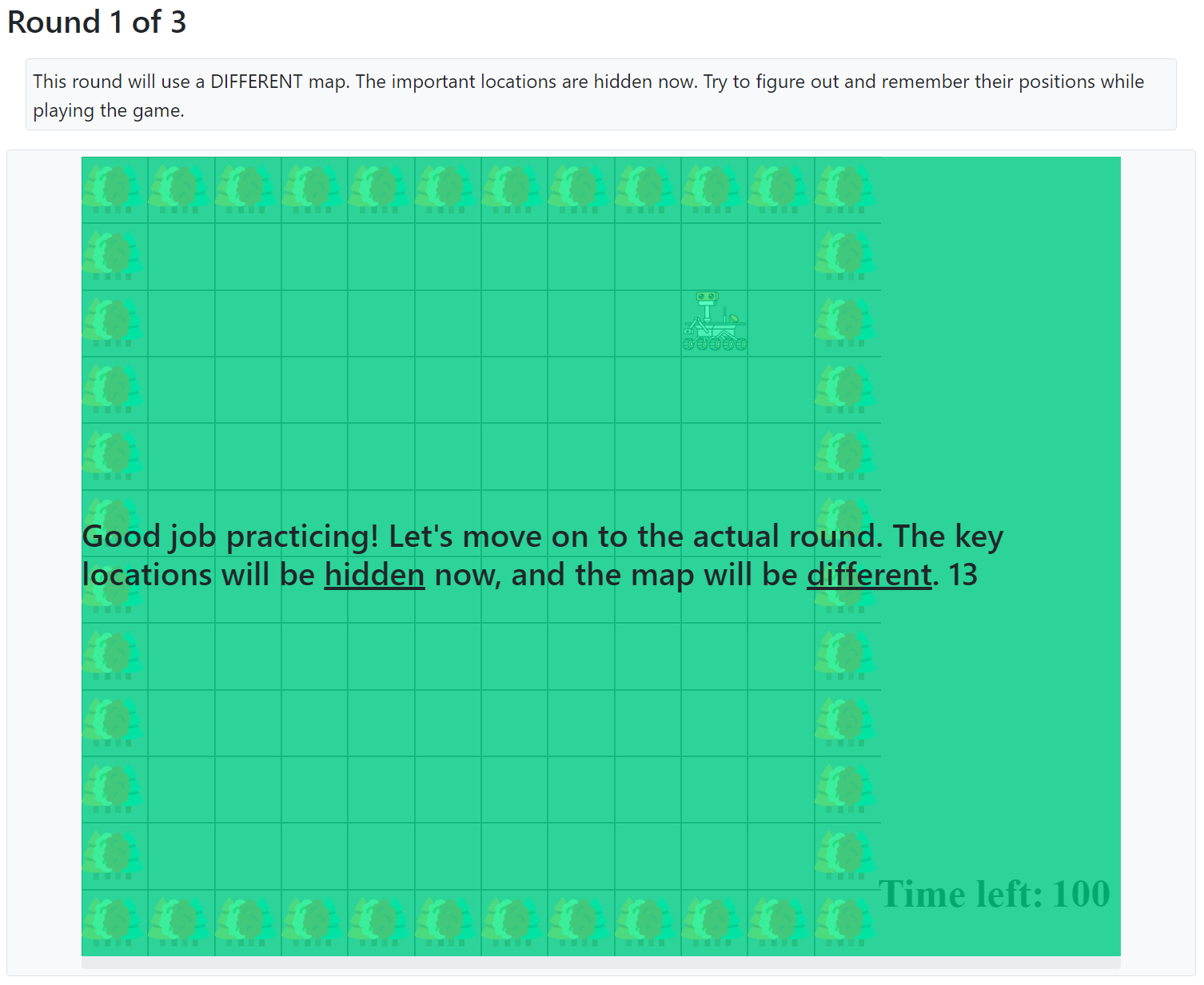}
        \label{fig:intrl_worker_9}
        \caption{After the trial round is over, the user is given instructions that the map will no longer display the locations of the game objects}
    \end{subfigure}
\end{figure*}
\begin{figure*}[h]
    \centering
    \begin{subfigure}
        \centering
        \includegraphics[width=2\columnwidth]{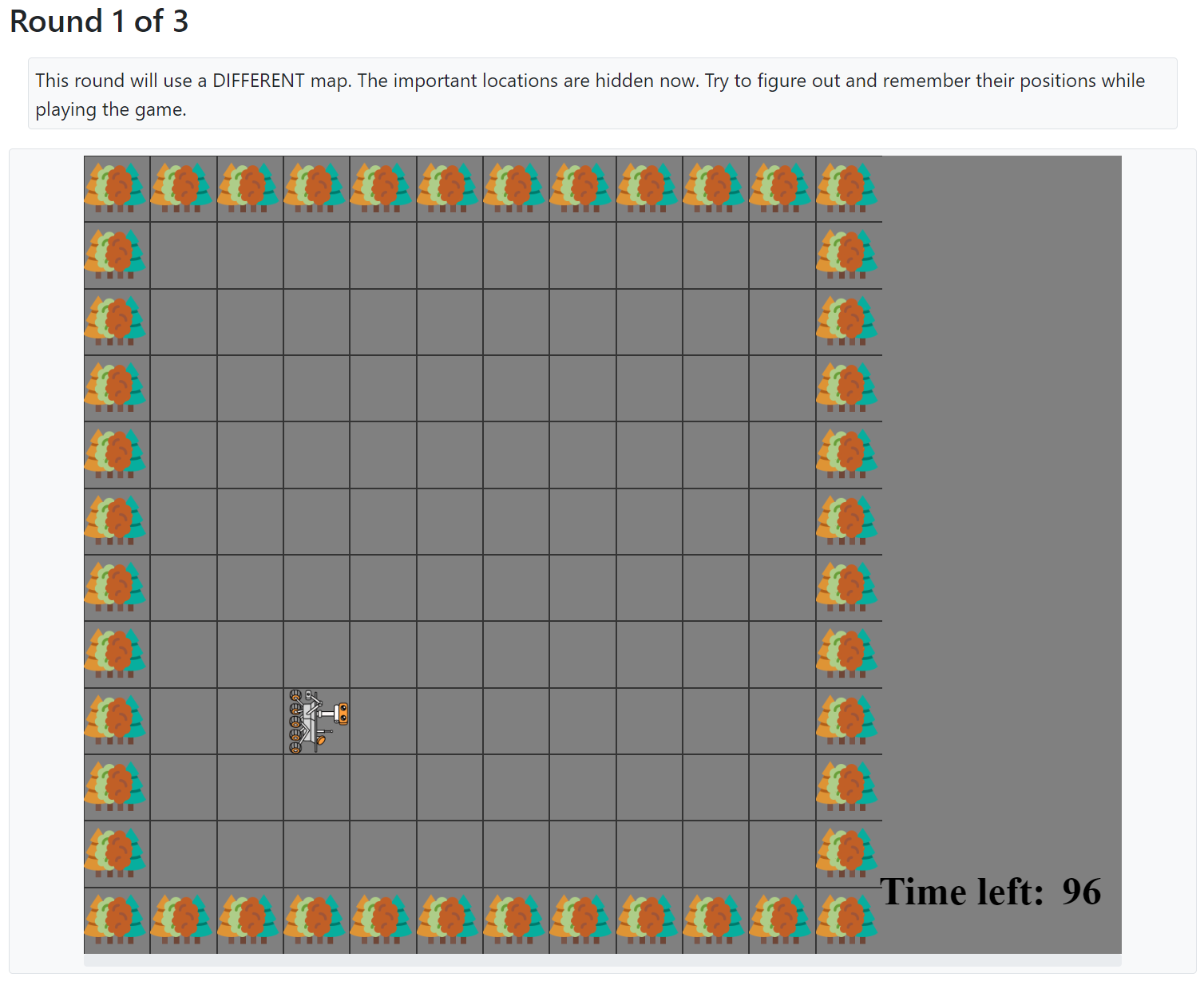}
        \label{fig:intrl_worker_10}
        \caption{In the main recording session, the worker is unable to see any game objects nor their location}
    \end{subfigure}
\end{figure*}
\begin{figure*}[h]
    \centering
    \begin{subfigure}
        \centering
        \includegraphics[width=2\columnwidth]{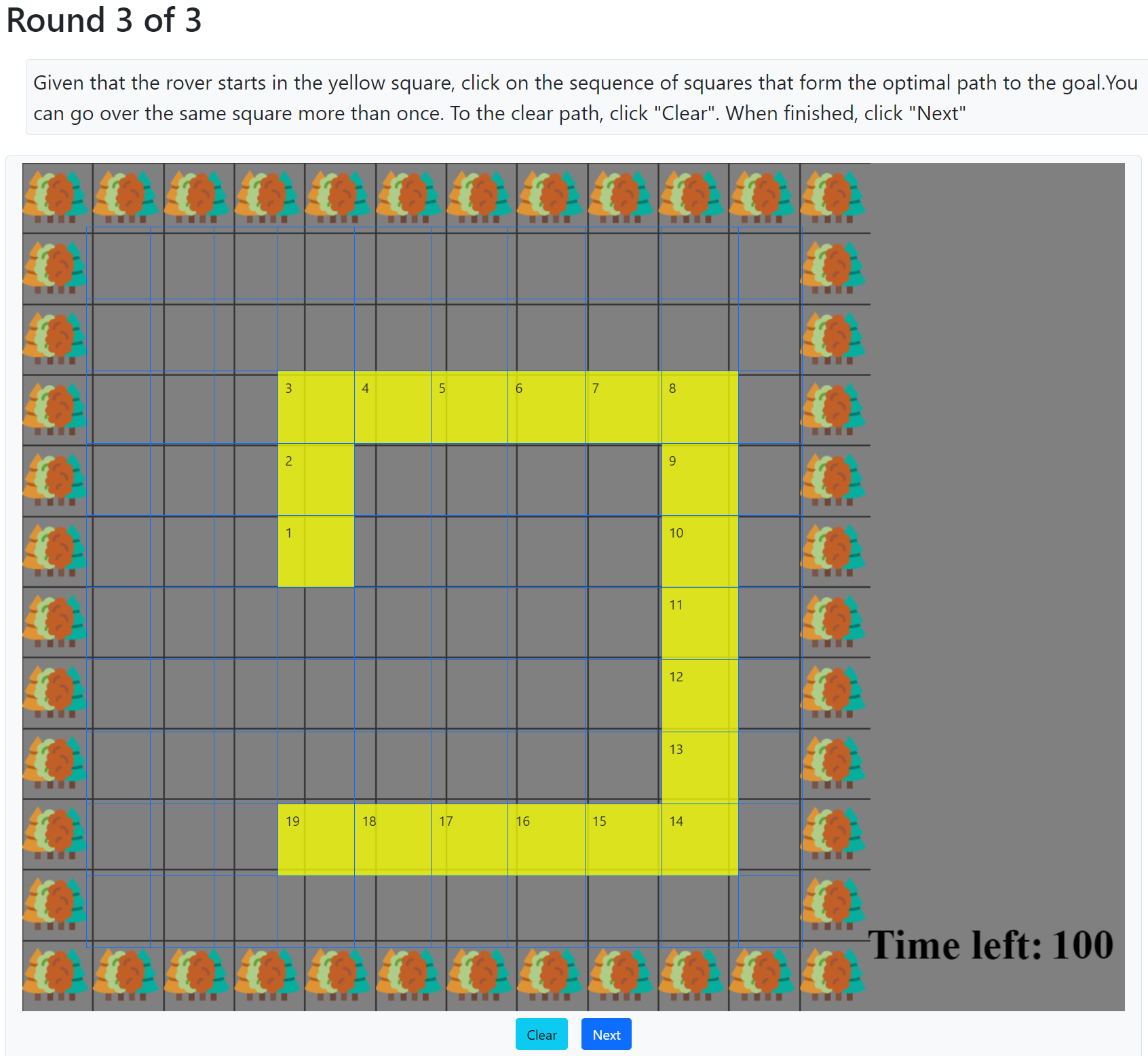}
        \label{fig:intrl_worker_11}
        \caption{After all of the game rounds complete, we ask the Worker to guess the optimal path for the rover}
    \end{subfigure}
\end{figure*}
\begin{figure*}[h]
    \centering
    \begin{subfigure}
        \centering
        \includegraphics[width=1.8\columnwidth]{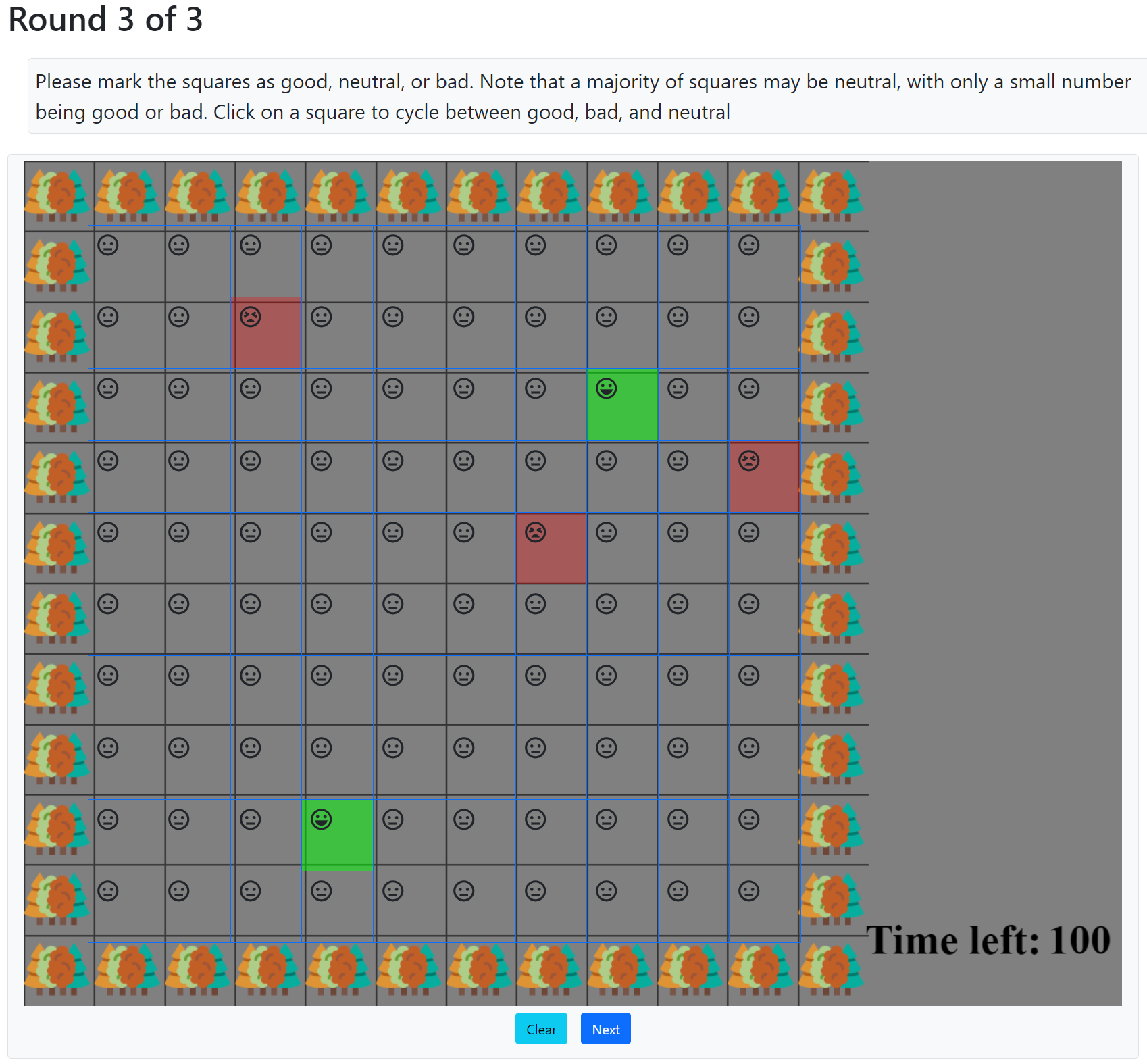}
        \label{fig:intrl_worker_12}
        \caption{Additionally, we prompt the Worker to annotate where the good and bad objects are located on the map}
    \end{subfigure}%
    \begin{subfigure}
        \centering
        \includegraphics[width=0.7\textwidth]{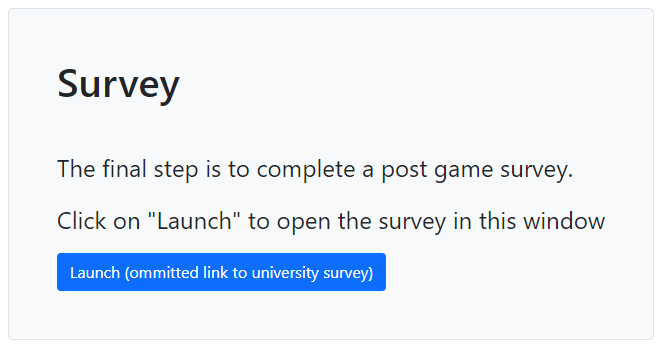}
        \label{fig:intrl_worker_13}
    \end{subfigure}
\end{figure*}

\clearpage
\subsection{TAMER Implementation}
\label{appendix-tamer}

The implementation followed the instructions by Knox and Stone \cite{knox-2009}. This included a discount factor of $\gamma = 0$. Therefore, the agent was completely myopic and only learned about actions that produce an immediate reward \cite{knox-2009}. Additionally, the MDP reward function \textit{R} was replaced by a human reward function \textit{H}, which was completely based on the feedback given by the human trainer. The human reward function was learned by a supervised learning model \cite{knox-2009}. A kernelized regression model was implemented with a kernel union to cover different gammas and variances. The library scikit-learn was used for the implementation of the regression models and the kernel featurizer. Moreover, the weights of the regression model were updated with a stochastic gradient descent algorithm and a learning rate of $0.01$.

Following the approach of Deep-Q learning \cite{sutton2018reinforcement}, the output of the human feedback model represents state-action values. We implemented four supervised models, one for each action, since regression models only give one output scalar. Each human feedback model took a vector representing the agent's current state as input. The state was the row and column of the agent's location as well as a bool whether the nut had been collected or not. The output value represented the state-action value for the input state, and the action represented by the supervised model. The agent learned \textit{H} by updating one specific action model once a human reward was given for one state-action pair. If no reward was obtained and $h = 0$, the model(s) were not updated. 

Furthermore, we implemented credit assignment as described by Knox and Stone \cite{knox-2009}. Credit assignment splits the reward over the \textit{n} most recent state-action pairs. As suggested by the authors, we implemented a probability density function (pdf) with gamma($2.0, 0.28$). With the help of the pdf, weights for each of the state-action pairs are calculated, which represent the probability that the reward was meant for that pair. The weights were calculated for each state-

\begin{minipage}{0.9\fulltextwidth}
    \begin{figure}[H]
        \centering
        \includegraphics[width=\textwidth]{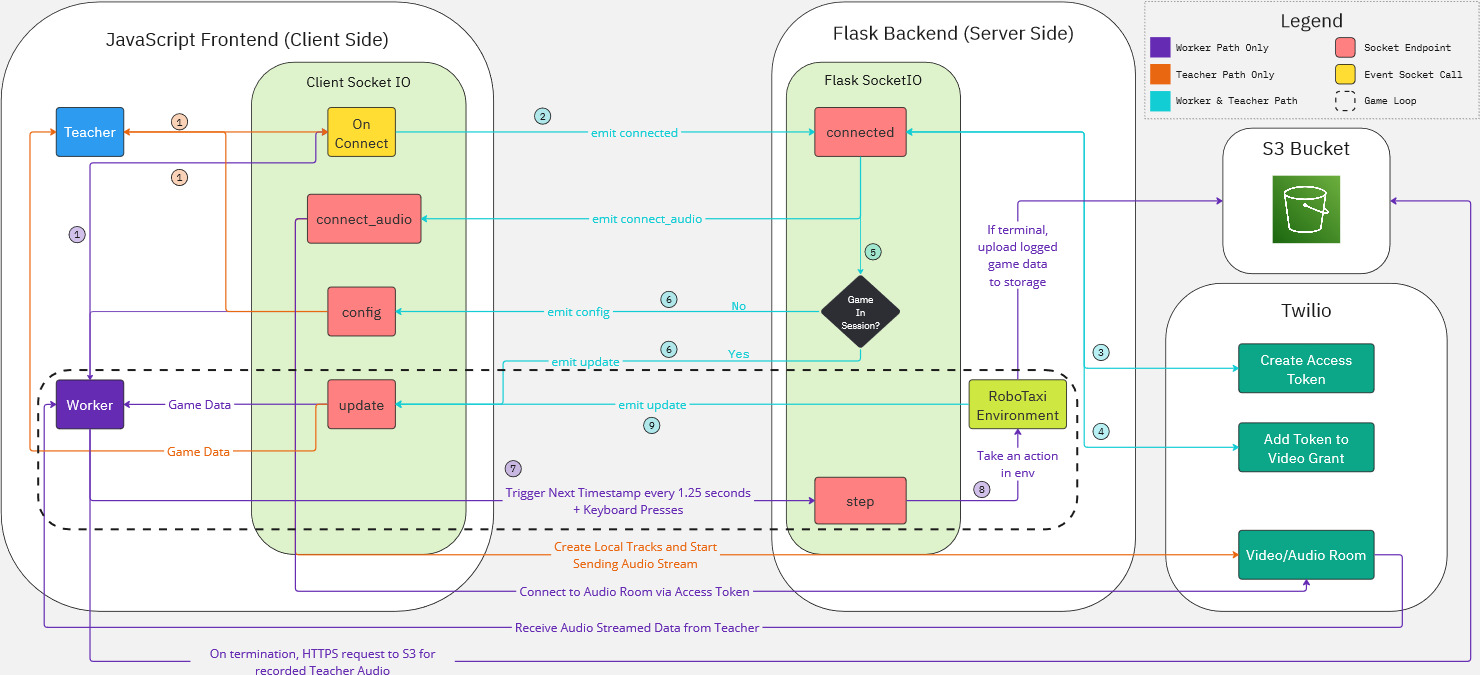}
        \caption{System Design for the Wizard of Oz data collection setup for the interactive RL setup.}
        \label{fig:intrl_socket_diagram}
    \end{figure}
\end{minipage}

\noindent action pair $t_{1}, t_{2},..., t_{n}$, where if $i < j$, then $i$ happened before $j$. Thus, $t_{n}$ was the most recent timestamp. The timestamp of when the human feedback was obtained was denoted with $0$ on the gamma pdf. Then, the weights were iteratively calculated for each state-action pair by taking the integral $\int_{t_{i-1}}^{t_{i}}f(x)dx$, where $t_{i}$ represents the timestamp when the agent executed the action. The resulting weight, hence the probability that the reward was meant for that specific state-action pair, is then multiplied by the obtained human reward and used to update the human reward function $H$. We chose $n = 3$ because of the relatively low speed of the robot. Therefore, it was quite unlikely that the feedback was meant for an older agent timestamp.

Since the agent learned with offline learning, it did not choose its own actions. Instead, \textit{H} was updated based on the recorded state-action pairs and voice feedback of the WoZ experiment.

\subsection{System Design for WoZ data collection pipeline}
\label{appendix-socket-diagram}

We show a schematic diagram (Fig.~\ref{fig:intrl_socket_diagram}) of the system design for the WoZ data collection setup described in Sec.~\ref{sec:woz}.

\clearpage

\section{Appendix: Audio-augmented demonstrations}
\label{appendix-atari}
\subsection{Additional Results}
\label{appendix:atari_results}

We show additional results from the analysis of the user study data for three different Atari games (Fig. 16-19).
\vspace{1mm}
\begin{figure}[!htb]
    \centering
    \includegraphics[width=0.40\textwidth]{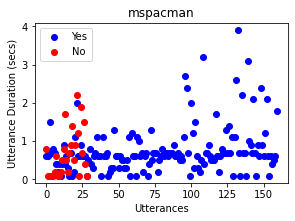}
    \includegraphics[width=0.40\textwidth]{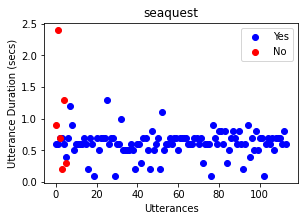}
    \includegraphics[width=0.40\textwidth]{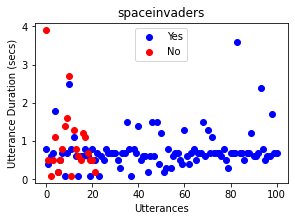}
    \caption{Duration (in seconds) of yes and no utterances from a single demonstrator, during demonstrations provided to three Atari games.}
\label{fig:utt_dur_atari}
\end{figure}

\begin{figure}[b]
    \centering
    \includegraphics[width=0.4\textwidth]{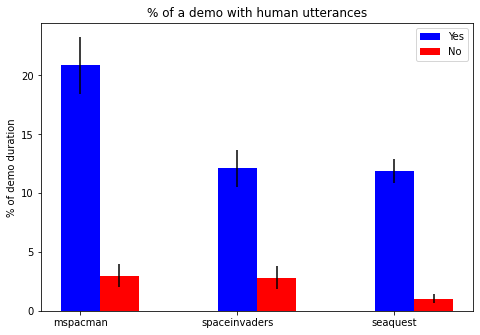}
    \caption{Percentage of yes and no utterance duration given the total demonstration duration.
    }
\label{fig:yes_no_percentage}
\end{figure}

\begin{figure}[b]
    \centering
    \includegraphics[width=0.4\textwidth]{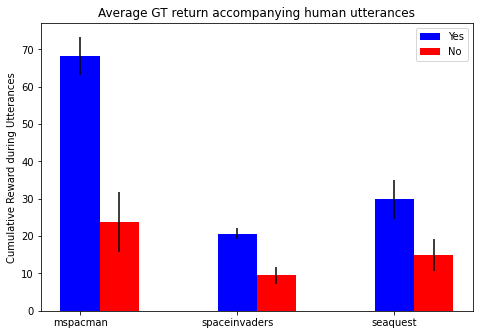}
    \caption{Cumulative reward of snippets surrounding audio utterances.
    }
\label{fig:yes_no_return}
\end{figure}

\begin{figure*}
\centering
\includegraphics[width=0.3\textwidth]{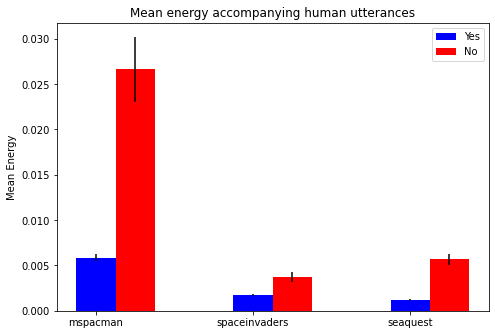}
\includegraphics[width=0.3\textwidth]{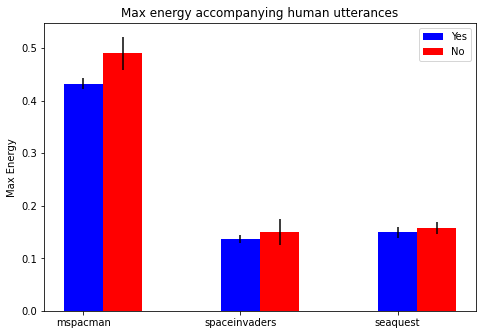}
\includegraphics[width=0.3\textwidth]{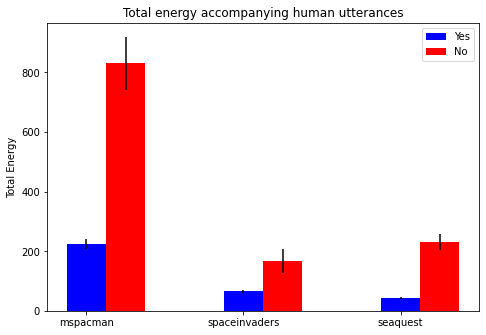}
\includegraphics[width=0.3\textwidth]{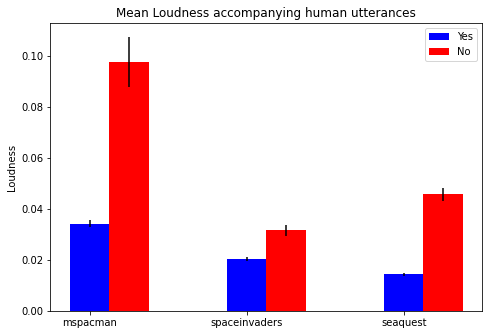}
\includegraphics[width=0.3\textwidth]{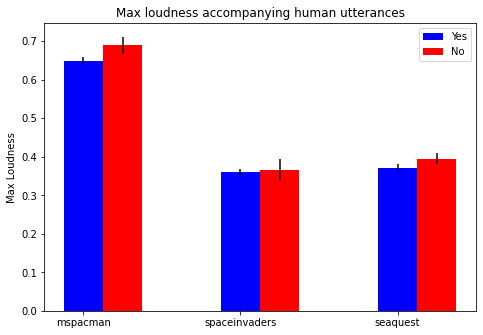}
\includegraphics[width=0.3\textwidth]{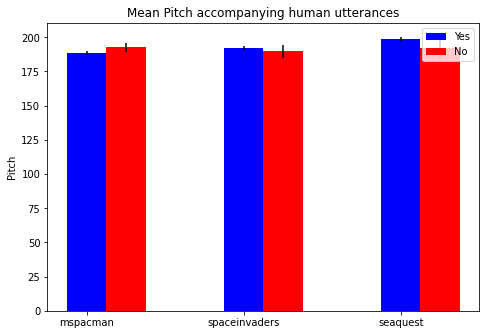}
\caption{Prosodic feature values for yes and no utterances from a single demonstrator providing demonstrations to three different Atari games. 
}
\label{fig:prosody_yes_no}
\end{figure*}

\clearpage

\subsection{Screenshots of Atari}
\label{appendix:screenshots_atari}

We share screenshots of the data collection interface provided to the demonstrator during a user study session for one of the three games (Fig. 20-21).

\begin{minipage}{\textwidth}
    \begin{figure}[H]
        \centering
        \begin{subfigure}
        \centering
            \includegraphics[height=6.1cm]{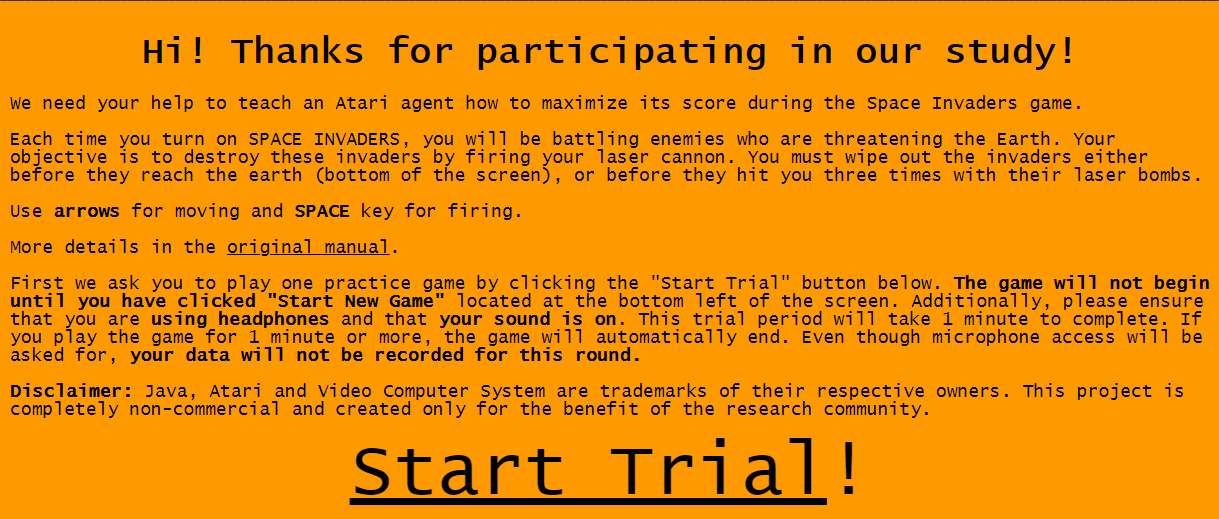}
            \label{fig:atari_screenshot_1}
            \vspace{0.2cm}
        \end{subfigure}
        \begin{subfigure}
            \centering
            \includegraphics[height=7.1cm]{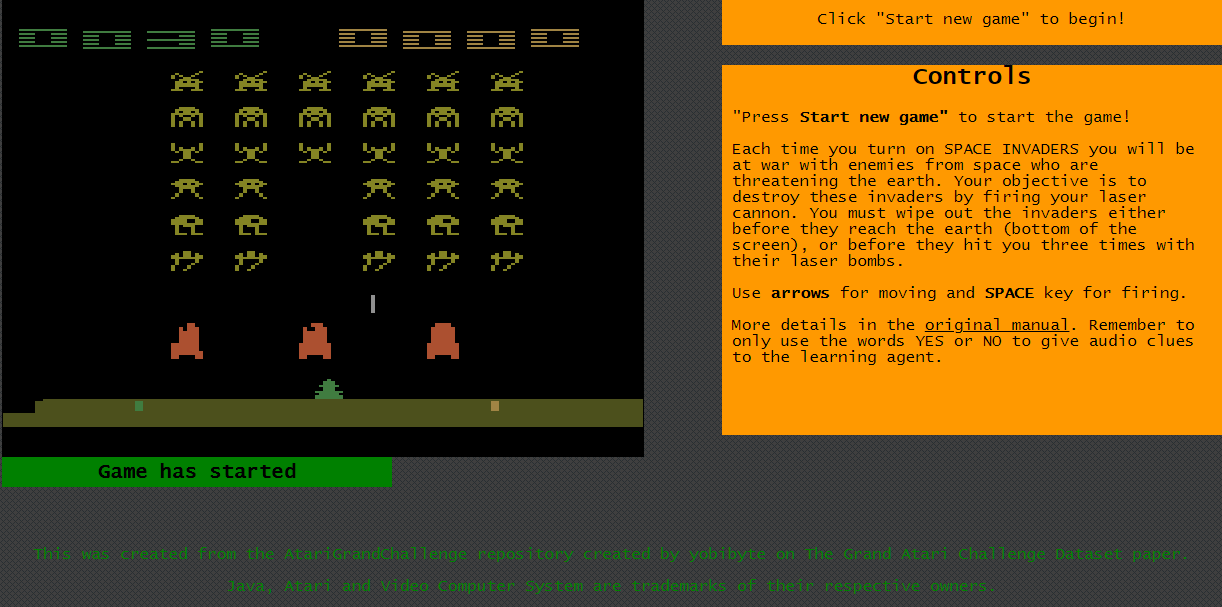}
            \label{fig:atari_screenshot_2}
        \end{subfigure}
        \caption{Trial period for the participant to get accustomed to the controls of the selected Atari game}
    \end{figure}
    \hspace{2cm}
\end{minipage}

\begin{figure*}[!h]
    \centering
    \begin{subfigure}
        \centering
        \includegraphics[height=6.0cm]{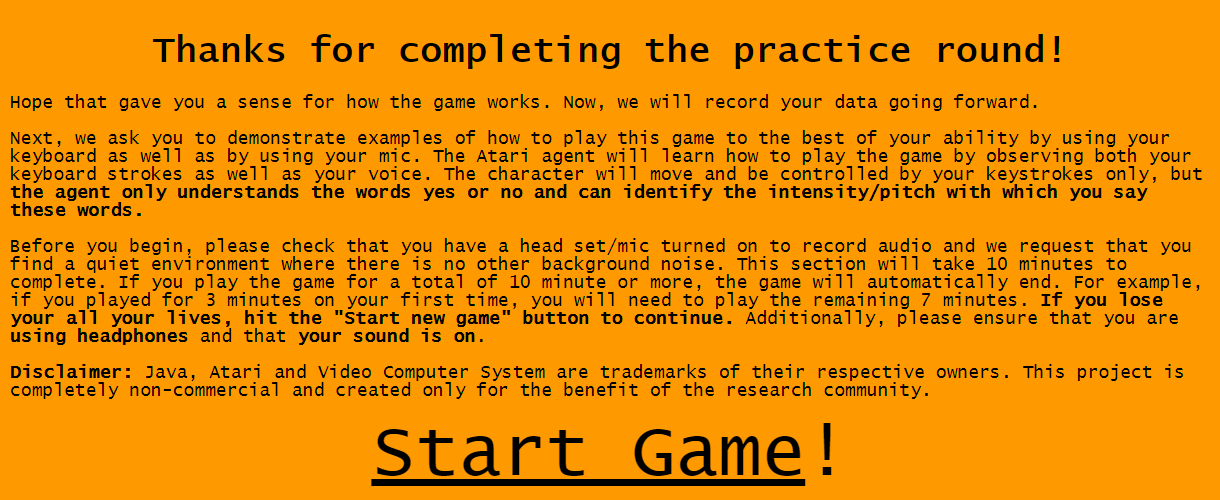}
        \label{fig:atari_screenshot_3}
        \vspace{0.2cm}
    \end{subfigure}
    \begin{subfigure}
        \centering
        \includegraphics[height=7.22cm]{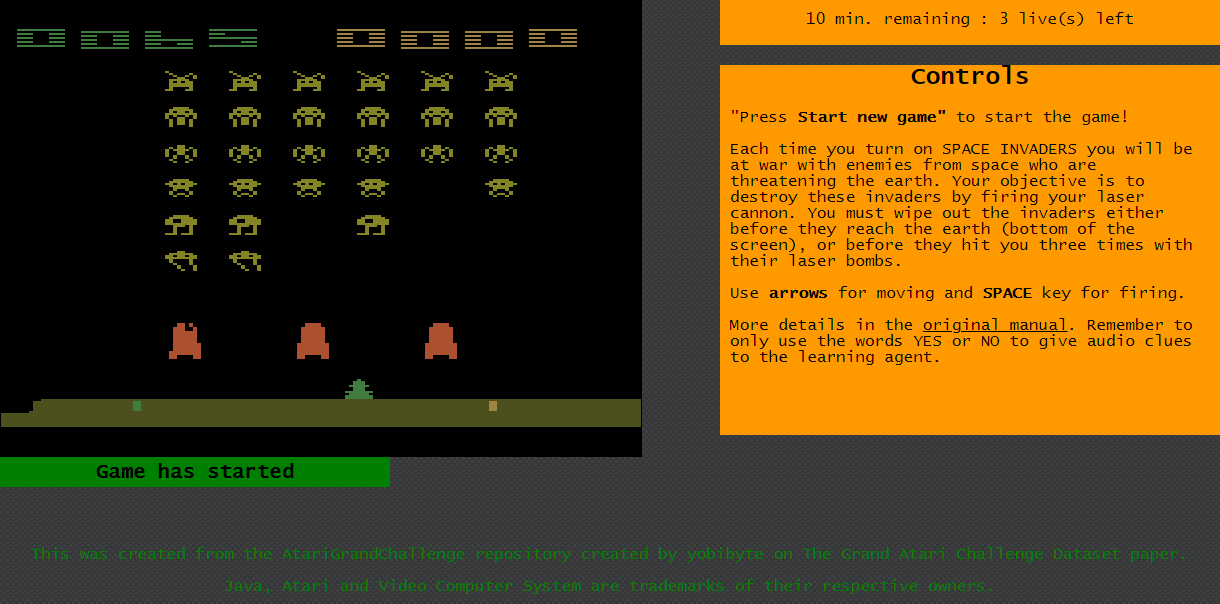}
        \label{fig:atari_screenshot_4}
        \vspace{0.2cm}
    \end{subfigure}
    \begin{subfigure}
        \centering
        \includegraphics[height=7.14cm]{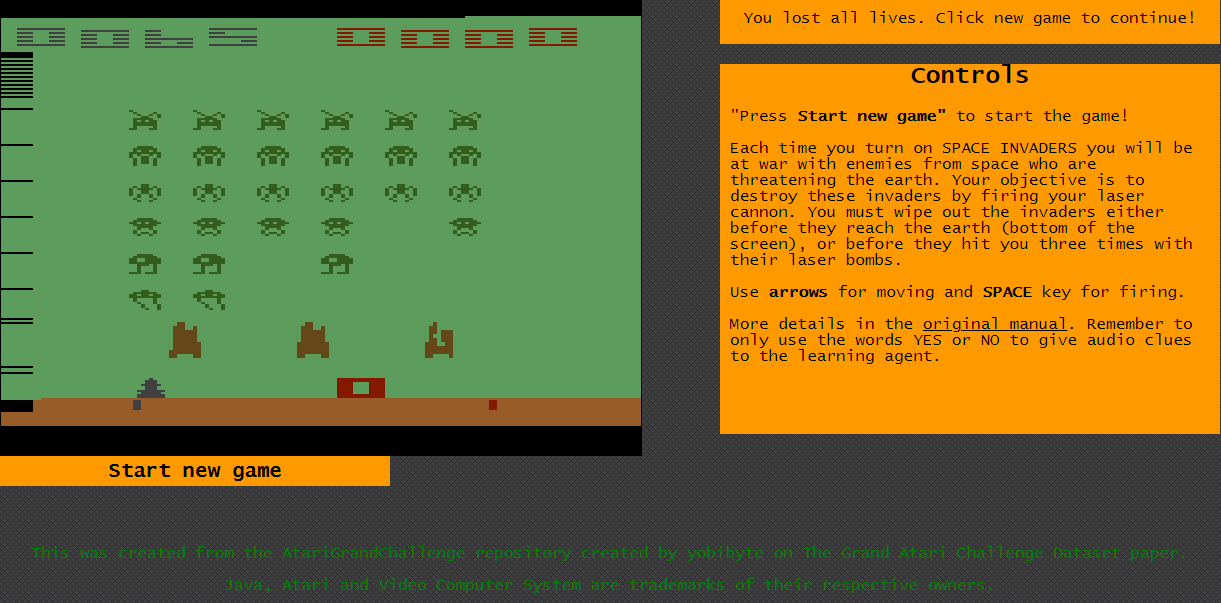}
        \label{fig:atari_screenshot_4_5}
        \vspace{0.2cm}
    \end{subfigure}
    \caption{Game period to record the participant's voice and game data as they play an Atari game.}
\end{figure*}

\newpage

\end{document}